\documentclass[11pt, a4paper, logo, copyright]{googledeepmind}

\usepackage[numbers, sort&compress]{natbib} 

\usepackage{hyperref}
\usepackage{xurl}
\usepackage{cleveref}
\usepackage{url}

\usepackage{wrapfig,booktabs}
\usepackage[table]{xcolor}        
\usepackage{tikz}
\usetikzlibrary{patterns}
\usepackage{colortbl}
\usepackage{arydshln}
\usepackage{adjustbox}
\usepackage{multirow}
\usepackage{graphicx}
\usepackage{makecell}              
\usepackage{booktabs}
\usepackage{threeparttable}        
\usepackage{sidecap}               

\usepackage{enumitem}

\usepackage{gensymb}
\usepackage{xspace}                
\usepackage{subcaption}

\usepackage{minitoc}               
\usepackage{makeidx}
\usepackage[toc,page,header]{appendix}

\usepackage{bbm}
\usepackage{amssymb}
\usepackage{amsmath}

\usepackage{algorithm}
\usepackage{algorithmic}

\usepackage{psfrag}
\usepackage{afterpage}
\usepackage{float}
\usepackage{array}
\usepackage{pifont}
\usepackage{placeins}              
\usepackage{titletoc}              

\usepackage{microtype}             
\usepackage{utfsym}                
\usepackage{orcidlink}             

\usepackage{caption}
\definecolor{sigterm}{RGB}{0, 102, 204}   
\definecolor{critterm}{RGB}{204, 0, 0}

\definecolor{lightgray}{gray}{0.9}

\colorlet{darkgreen}{green!65!black}
\colorlet{darkblue}{blue!75!black}
\colorlet{darkred}{red!80!black}
\definecolor{lightblue}{HTML}{0071bc}
\definecolor{lightgreen}{HTML}{39b54a}
\definecolor{manyshot}{HTML}{6969ff}
\definecolor{medshot}{HTML}{f7c600}
\definecolor{fewshot}{HTML}{ff6969}
\definecolor{mypurple}{HTML}{412F8A}
\definecolor{myorange}{HTML}{fc8e62}

\definecolor{deemph}{gray}{0.55}

\definecolor{textgreen}{RGB}{57, 172, 57}
\definecolor{textred}{RGB}{200, 10, 10}
\definecolor{textgray}{RGB}{100, 100, 100}
\definecolor{visiongold}{RGB}{230, 184, 0}
\definecolor{speechpurple}{RGB}{204, 0, 255}
\definecolor{dataprep}{RGB}{38, 189, 128}
\definecolor{modeltraining}{RGB}{38, 189, 128}
\definecolor{backgroundcol}{RGB}{232, 230, 230}
\definecolor{gold}{rgb}{225, 215, 200}     
\definecolor{navyblue}{RGB}{40, 66, 200}   
\definecolor{orange}{RGB}{255,127,80}
\definecolor{pink}{RGB}{219,112,147}

\definecolor{ForestGreen}{RGB}{0,235,110}
\definecolor{backred}{RGB}{255, 190, 190}
\definecolor{backblue}{RGB}{220, 230, 250}

\usepackage[most]{tcolorbox}
\tcbuselibrary{breakable,skins}
\usepackage{soul}
\usepackage{caption}
\usepackage{utfsym}
\usepackage{bbm}
\usepackage{algorithm}
\usepackage{algorithmic}
\usepackage{colortbl}
\usepackage{psfrag}
\usepackage{adjustbox}
\usepackage{wrapfig}
\usepackage{xspace}
\usepackage{amssymb}
\usepackage{multirow}
\usepackage{afterpage}
\usepackage{float}
\usepackage[toc,page,header]{appendix}
\usepackage{minitoc}
\usepackage{algorithm}
\usepackage{algorithmic}
\usepackage{array}
\usepackage{amsmath}
\usepackage{pifont}
\usepackage{enumitem}

\definecolor{sigterm}{RGB}{0, 102, 204}  
\definecolor{critterm}{RGB}{204, 0, 0} 

\definecolor{baselinecolor}{gray}{.95}

\newcommand{\graydashedline}{\arrayrulecolor{gray!80}\hdashline\arrayrulecolor{black}}

\newlist{myenum}{enumerate}{1}
\setlist[myenum]{leftmargin=2em,itemsep=0pt,topsep=0pt,partopsep=0pt}

\usepackage{makecell, tabularx}
\newcolumntype{L}{>{\RaggedRight}X}
\usepackage{siunitx}

\title{LiveK12Bench: Have Large Multimodal Models Truly Conquered High School-level Examinations?}

\correspondingauthor{shawnbywang@tencent.com; goodli@tencent.com}

\reportnumber{}

\renewcommand{\today}

\author[1,$\dagger$,*]{Xiaohan Wang}
\author[1,2,*]{Mingze Yin}
\author[1]{Yilin Zhao}
\author[1]{Gang Liu}
\author[1,$\dagger$]{Dian Li}

\affil[1]{Tencent PCG}
\affil[2]{College of Computer Science and Technology, Zhejiang University}
\affil[$\dagger$]{Corresponding Author}
\affil[*]{Equal Contribution}

\begin{abstract}
Advanced Large Multimodal Models (LMMs) have demonstrated impressive performance in K-12 reasoning tasks, exhibiting great promise as intelligent tutors. Realizing this potential requires models to navigate real-world examinations effectively, yet most existing benchmarks fail to capture the complexity of authentic testing environments. Specifically, most datasets are static, prone to data contamination, and are often confined to restricted modalities, disciplines, and evaluation criteria. To address these issues, we introduce \textbf{LiveK12Bench}, a dynamic, holistic, multi-disciplinary benchmark designed to evaluate the reasoning abilities of LMMs in realistic examination scenarios. LiveK12Bench comprises 2K+ verified questions spanning Mathematics, Physics, Chemistry, and Biology, sourced from the latest real-world exam papers and designed to grow over time. Our framework features several core innovations: 1) featuring an automated pipeline that continuously ingests and parses latest examination papers to mitigate data leakage; and 2) proposing a novel `Mock Exam' evaluation scheme, which assesses the ability to complete end-to-end exams autonomously with accurate and efficient reasoning paths. Extensive experiments on 12 LMMs reveal that advanced models suffer substantial performance degradation under exam-realistic constraints: GPT-5's score drops from 79 to 53 (out of 100) when process rigor and efficiency are jointly evaluated. Our findings expose critical vulnerabilities, such as sensitivity to complex visual layouts, highlighting the gap between idealized reasoning capabilities and true educational readiness. Both \href{https://github.com/QQ-MM/LiveK12Bench}{code} and 
\href{https://huggingface.co/datasets/Shawn-wxh/livek12bench}{dataset} are publicly available.
\end{abstract}

\begin{document}

\maketitle

\newenvironment{Itemize}{
    \begin{itemize}[leftmargin=*]
    \setlength{\itemsep}{0pt}
    \setlength{\topsep}{0pt}
    \setlength{\partopsep}{0pt}
    \setlength{\parskip}{0pt}}
{\end{itemize}}
\setlength{\leftmargini}{9pt}

\section{Introduction}

Generative AI is rapidly transforming the educational landscape. As large language models continue to push the boundaries of their reasoning capabilities, they have achieved near-perfect performance on high school-level mathematical benchmarks such as MATH \cite{hendrycks2021measuring} and AIME \cite{aime25,ye2025aimepreview}. However, to truly serve as effective and reliable tutors for human students, AI must first demonstrate the ability to successfully navigate authentic human examinations. While recent news frequently highlight that advanced LMMs can achieve impressive scores on college entrance examinations, a critical question remains: \textit{Have large multimodal models truly conquered high school-level examinations?}

To drive the evolution of AI reasoning, mainstream research has predominantly focused on highly logic-dependent disciplines such as mathematics and programming. To isolate reasoning capabilities from other confounding factors, traditional
benchmarks typically provide meticulously parsed question information and are confined to evaluating easily verifiable final answers. Recently, some benchmarks have shifted their focus toward evaluating generative AI in educational contexts,
introducing K-12 multi-disciplinary assessments and photo-based problem-solving evaluations \cite{zhang2023m3exam, ye2025mmscibench,das2024exams}. However, these existing benchmarks struggle to comprehensively answer the aforementioned question, as they fundamentally fail to bridge three core gaps between AI evaluation and authentic human testing: 1) \textbf{Data Leakage}: Most datasets are static. Once published, they are inevitably ingested into the training corpora of next-generation LLMs, rendering subsequent evaluations unreliable and losing their reference value \cite{sainz2023nlp}. 2) \textbf{Insufficient Evaluation}: Human examinations assess students under

\begin{wrapfigure}[16]{L}[0pt]{0.5\textwidth}
    \centering
    \includegraphics[width=\linewidth]{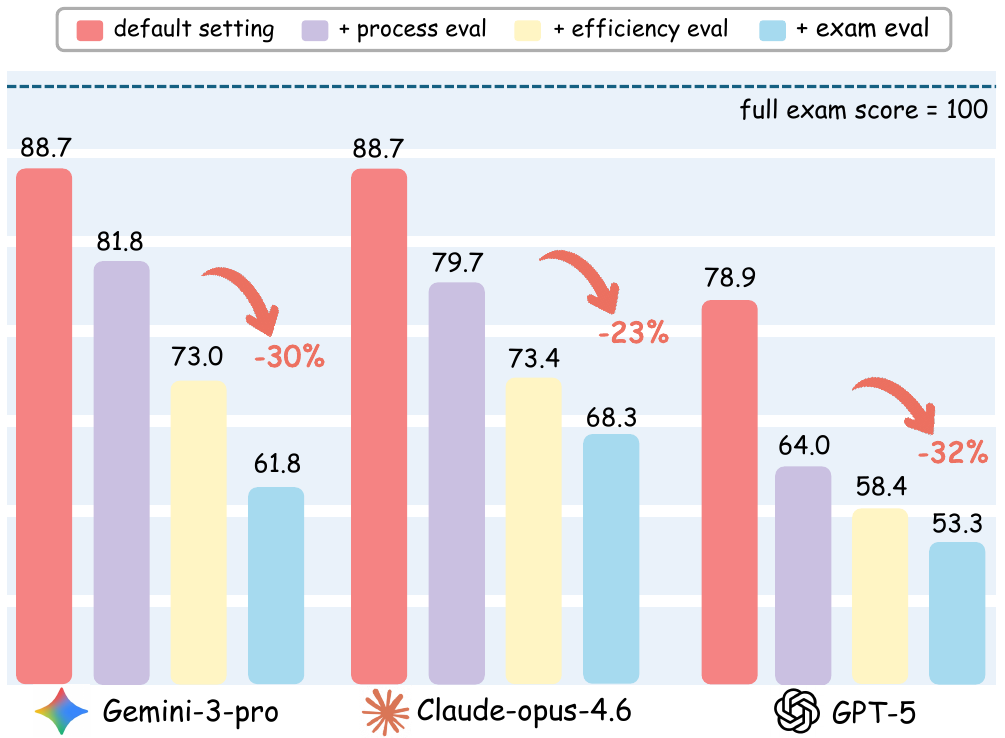}
    \caption{\textbf{Performance degradation of cutting-edge LMMs in authentic exam scenarios.}}
    \label{figure1}
\end{wrapfigure}

\noindent strict time and environmental constraints, comprehensively grading both the final answer and the step-by-step reasoning process across questions of varying importance. In contrast, AI evaluation criteria remain largely one-dimensional. 3) \textbf{Human Intervention}: Testing models on K-12 exam papers often involves manual question extraction, image cropping, or providing textual descriptions for visual elements. Consequently, the task input for AI is not equivalent to that of a human student, precluding a genuine end-to-end examination. These gaps make it exceedingly difficult to accurately estimate the practical usability and potential value of mainstream LMMs as intelligent tutors or educational assistants.

To address these limitations, we propose LiveK12Bench, a dynamic, comprehensive, and multi-disciplinary AI examination benchmark designed to systematically investigate the capabilities and limitations of mainstream LMMs in real-world K-12 scenarios. Specifically, to eradicate the issue of test data leakage at its source (and to avoid the pitfalls of AI-generated synthetic questions), LiveK12Bench introduces a highly efficient, automated examination paper parsing pipeline based on structural document extraction and LLM parsing. This pipeline enables the periodic ingestion of fresh questions newly authored by frontline educators, continuously expanding the dataset scale. Concurrently, we propose a ``Mock Exam'' evaluation scheme that simulates the multi-dimensional assessment of human exams, evaluating mainstream models across four dimensions: answer accuracy, process correctness, reasoning efficiency, and a weighted comprehensive exam score. Building upon standard text-only and text-image multimodal settings, we introduce an ``Image-Only'' full-page modality. This setting aligns with an end-to-end testing scenario, significantly reducing manual assistance and intervention.

By evaluating mainstream multimodal reasoning models (illustrated in Figure \ref{figure1}), the performance of leading LMMs degrades across three progressively challenging scenarios: from standard settings to exam scoring incorporating process and efficiency evaluation, and end-to-end ``Image-Only'' exam modality. These analytical insights provide crucial implications for the future development of generative AI in educational applications.

Our primary contributions are summarized as follows:
\vspace{-0.3cm}
\begin{itemize}
    \item[$\bullet$] We propose the first comprehensive, multi-disciplinary benchmark that holistically simulates authentic human K-12 examinations.
    \item[$\bullet$] We design an automated exam paper ingestion pipeline that facilitates the efficient and dynamic iteration of the dataset, effectively mitigating data contamination.
    \item[$\bullet$] We introduce a comprehensive ``Mock Exam'' evaluation protocol assessing reasoning process and problem-solving efficiency, and incorporating an test paper ``Image-Only'' input modality to impose real-world layout noise.
\end{itemize}

\section{LiveK12Bench}
\begin{figure*}[!t]
    \centering
    \includegraphics[width=\textwidth]{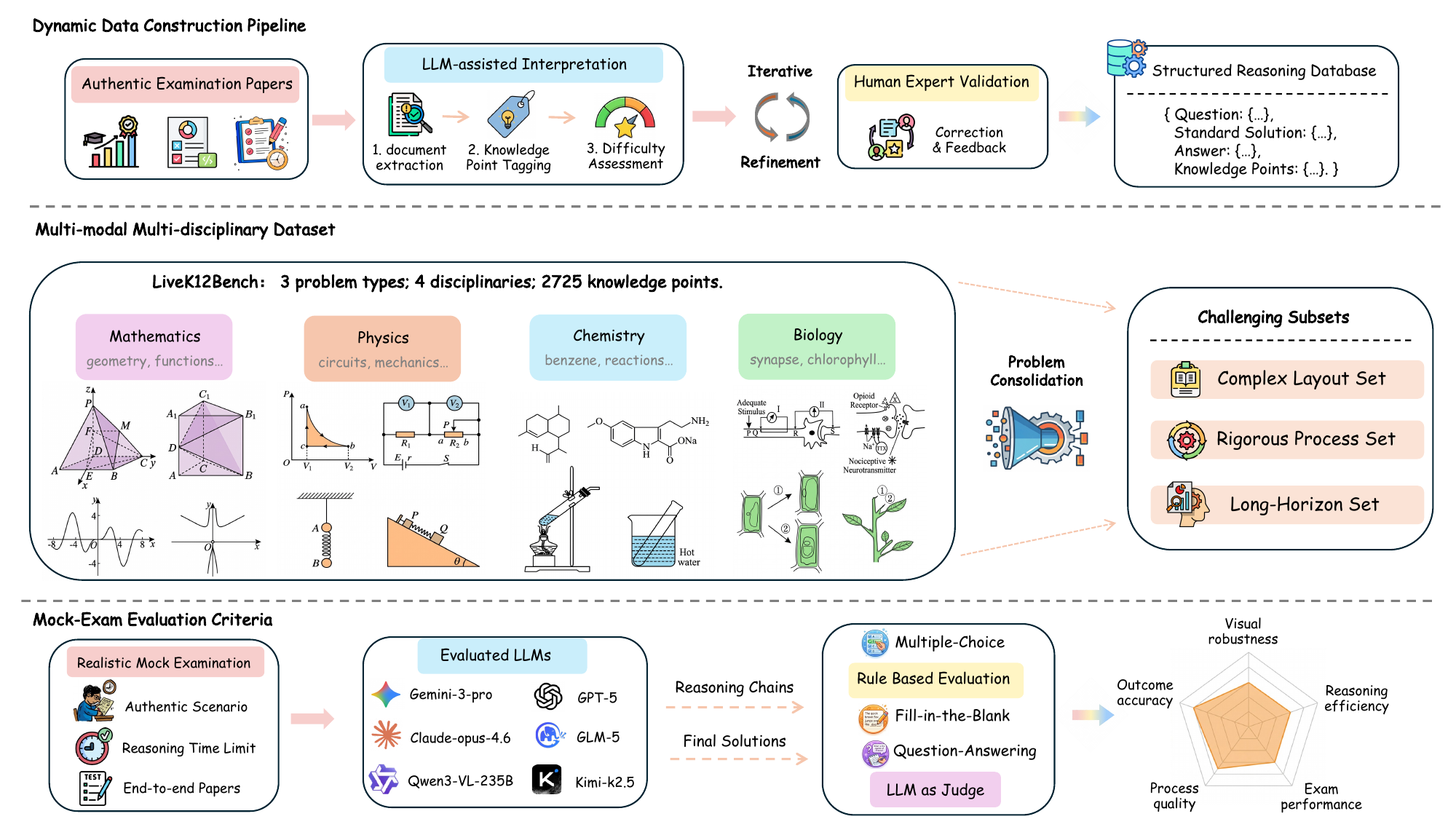}
    \caption{\textbf{Overall framework of LiveK12Bench.} The framework consists of three interconnected modules that conduct multi-dimensional evaluation starting from raw exam papers: a dynamic data construction pipeline, a comprehensive dataset, and a Mock-Exam evaluation protocol.}
    \label{fig:architecture}
\end{figure*}

LiveK12Bench systematically resolves the aforementioned challenges through three core innovations: a holistic dataset encompassing multi-disciplinary and multi-modal scenarios, a dynamic data construction pipeline that continually ingests fresh examination papers, and a novel ``Mock-Exam'' evaluation protocol that demands end-to-end problem solving with process, efficiency and modality constraints. The overall architecture and workflow of our framework are illustrated in Figure \ref{fig:architecture}. The following subsections detail the design and implementation of each component.

\subsection{Dataset Composition and Comparison}

The LiveK12Bench dataset currently comprises 2,114 high-quality, manually verified questions covering four core K-12 disciplines that heavily rely on reasoning capabilities: Mathematics, Physics, Chemistry, and Biology. The dataset encompasses diverse question formats, including Multiple-Choice Questions (MCQs), Fill-in-the-Blank (FIB) questions, and Q\&A questions. The dataset consists of two timestamp splits of 26-03 and 26-05 (indicating the release time of questions), both with Chinese and English translation versions.

To comprehensively evaluate the robustness of Large Language Models (LLMs) and Large Multimodal Models (LMMs), we curate tasks across three distinct modalities, corresponding to three realistic evaluation scenarios. Formally, let $f_\theta$ denote the evaluated model, $\mathcal{T}$ denote textual input, and $\mathcal{V}$ denote visual input. The three task modalities are defined as follows:
\vspace{-0.2cm}

\begin{itemize}
    \item \textbf{Text-Only (TO):} Both the input and expected output are purely textual, evaluating the foundational linguistic and symbolic reasoning capabilities of LLMs. The task is formulated as $A = f_\theta(\mathcal{T}_{q})$, where $\mathcal{T}_{q}$ represents the textual question stem and options, and $A$ is the textual predicted answer.
    \item \textbf{Text-Image (TI):} The input consists of an interleaved mixture of text and images, assessing the LMM's ability to ground textual concepts in visual representations (\textit{e.g.}, geometric diagrams, circuit schematics, or biological structures). The task is defined as $A = f_\theta(\mathcal{T}_{q}, \mathcal{V}_{q})$, where $\mathcal{V}_{q}$ denotes the cropped image(s) essential for solving the problem.
    \item \textbf{Image-Only (IO, Exam mode):} The input is an uncropped snapshot of a full examination page alongside a target question index, simulating the authentic task environment of a human student. It intentionally removes human-assisted intermediate steps, such as manual OCR and image cropping. The model must autonomously locate, extract, and interpret the relevant problem information from the page layout. The task is formalized as:
    \begin{equation}
        A = f_\theta(\mathcal{V}_{pages}, idx)
    \end{equation}
    where $\mathcal{V}_{page}$ is the raw exam page images, and $idx$ is the specified question number to be solved.
\end{itemize}

Beyond the core inputs, the dataset provides rich metadata for each question, including the ground-truth final answer annotated by professional educators, step-by-step solution paths, question types, score value, subject categories, and fine-grained knowledge point tags.

\begin{table*}[!t]
\centering
\caption{\textbf{Key statistics of LiveK12Bench.}}
\label{tab:dataset_statistics}
\resizebox{0.98\textwidth}{!}{%
\setlength{\tabcolsep}{3.5pt}
\begin{tabular}{lccccc}
\toprule
\textbf{Category} & \textbf{Overall}
& \raisebox{-0.85em}{\includegraphics[height=2.2em]{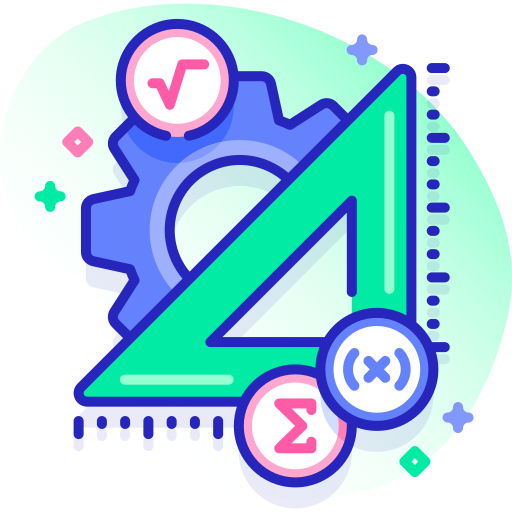}} \textbf{Mathematics}
& \raisebox{-0.7em}{\includegraphics[height=2em]{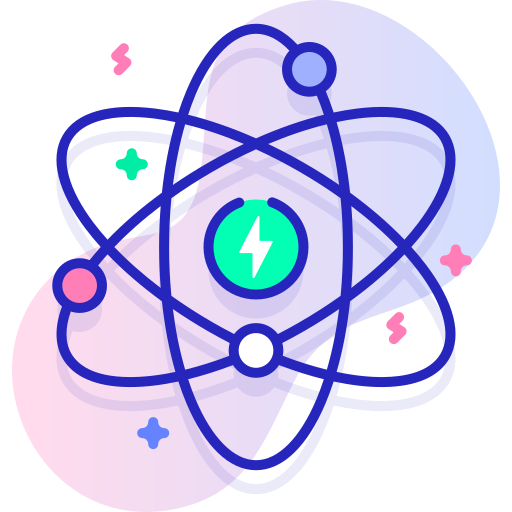}} \textbf{Physics}
& \raisebox{-0.7em}{\includegraphics[height=2em]{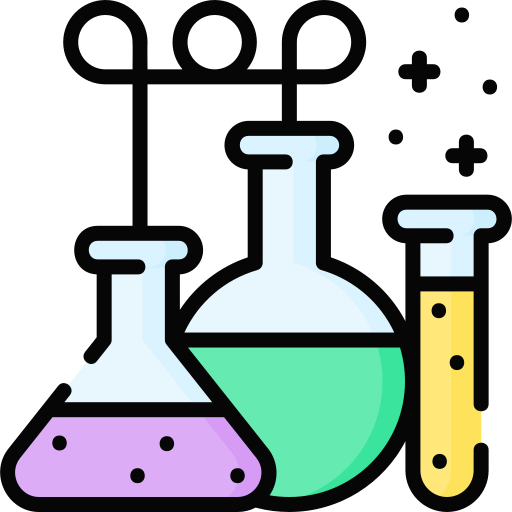}} \textbf{Chemistry}
& \raisebox{-0.85em}{\includegraphics[height=2.2em]{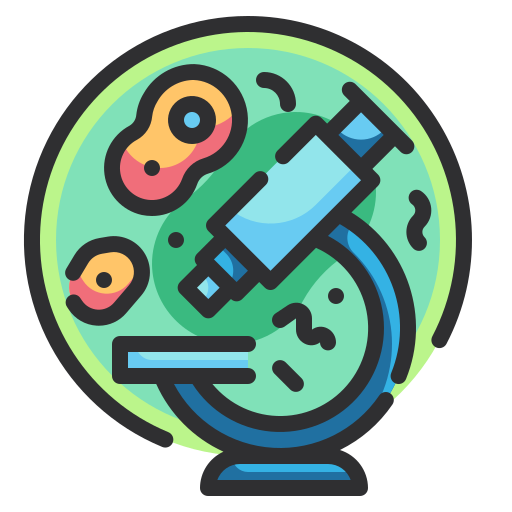}} \textbf{Biology} \\
\midrule
\multicolumn{6}{l}{\textit{Task Modality}} \\
- Text-Only (TO) & 1,096 & 617 (56.3\%) & 65 (5.9\%) & 240 (21.9\%) & 174 (15.9\%) \\
- Text-Image (TI) & 1,018 & 155 (15.2\%) & 331 (32.5\%) & 292 (28.7\%) & 240 (23.6\%) \\
- Image-Only (IO, Exam-mode) & 2114& 772 (36.5\%)& 220 (10.4\%)& 532 (25.2\%)& 414 (19.6\%) \\
\midrule
\multicolumn{6}{l}{\textit{Question Type}} \\
- Multiple-Choice (MCQ) & 1,473 & 444 (30.1\%) & 274 (18.6\%) & 419 (28.4\%) & 336 (22.8\%) \\
- Fill-in-the-Blank (FIB) & 164 & 119 (72.6\%) & 26 (15.9\%) & 18 (11.0\%) & 1 (0.6\%) \\
- Question-Answering (Q\&A) & 477 & 209 (43.8\%) & 96 (20.1\%) & 95 (19.9\%) & 77 (16.1\%) \\
\midrule
\textbf{Total Number} & \textbf{2,114} & \textbf{772 (36.5\%)} & \textbf{396 (18.7\%)} & \textbf{532 (25.2\%)} & \textbf{414 (19.6\%)} \\

\bottomrule
\end{tabular}%
}
\end{table*}

To probe the unique vulnerabilities of models, prior benchmarks\cite{MathVista,We-Math2.0} have constructed multiple subsets assessing different aspects of visual reasoning, such as measurement and puzzle tests. In line with this approach, aiming to capture the unique challenges of real-world K-12 examinations, we deliberately establish three subsets (50 questions per subject per subset, 600 in total) from our benchmark as follows:
\vspace{-0.3cm}
\begin{enumerate}
    \item \textbf{Complex Layout Set:} This subset specifically targets the visual challenge of real-world test papers, featuring question layouts with highly complex visual formatting. It includes cases where problems span across multiple pages, question stems are spatially detached from their corresponding figures, or images are tightly embedded within text blocks. This set challenges the LMM's capacity to accurately extract reasoning contexts from noisy visual margins and complex layouts (\textit{e.g.}, interpreting function curves and data tables). End-to-end proficiency on this set is a prerequisite for deploying AI educators in real-world, unconstrained input environments.
    
    \item \textbf{Rigorous Process Set:} Mainstream benchmarks predominantly evaluate final answer accuracy. However, the design of MCQs in human exams often allows students to guess the correct option through elimination or surface-level heuristics without rigorously deducing the underlying concepts. We specifically compile questions that are assigned with multiple knowledge points and are susceptible to such ``lucky guesses'' with the feature of excessive premises. This set is designed to evaluate the model's ability to arrive at the correct answer through a logically sound and solid reasoning process (process evaluation methodology is detailed in Section \ref{sec:mock_exam_evaluation}).
    
    \item \textbf{Long-Horizon Reasoning Set:} This subset comprises problems that frequently trap models in excessively long or circular reasoning chains, where the questions typically pose complex objectives and are assigned with higher max points in the papers. The difficulty may stem from intrinsic mathematical complexity, intentionally confounding conditions, or deceptive visual information. It aims to specifically evaluate the \textit{reasoning efficiency} of LMMs. Intuitively, an AI model that requires fewer generated tokens to correctly solve a complex problem offers superior computational efficiency and user experience.
    \label{long_reason}
\end{enumerate}

\vspace{-0.2cm}

\begin{figure*}[!t]
    \centering
    \includegraphics[width=\textwidth]{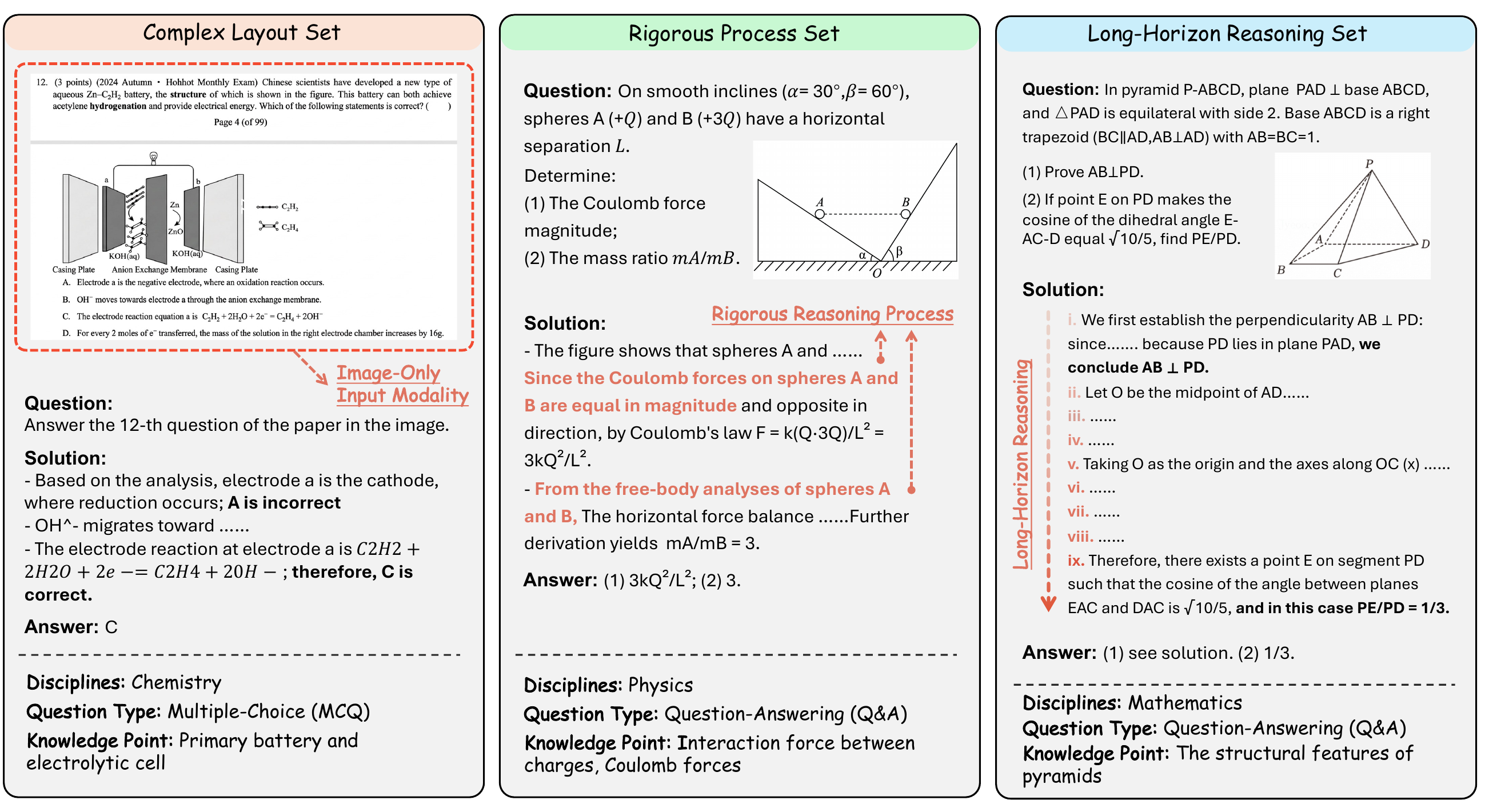}
    \caption{\textbf{Examples from the Three Challenging Subsets in LiveK12Bench.} The figure illustrates typical inputs and annotated metadata for the Complex Layout Set, Rigorous Process Set, and Long-Horizon Reasoning Set.}
    \vspace{-0.3cm}
    \label{fig:challenge_subsets}
\end{figure*}

\begin{wrapfigure}[17]{l}{0.5\textwidth}
    \centering
    \includegraphics[width=\linewidth]{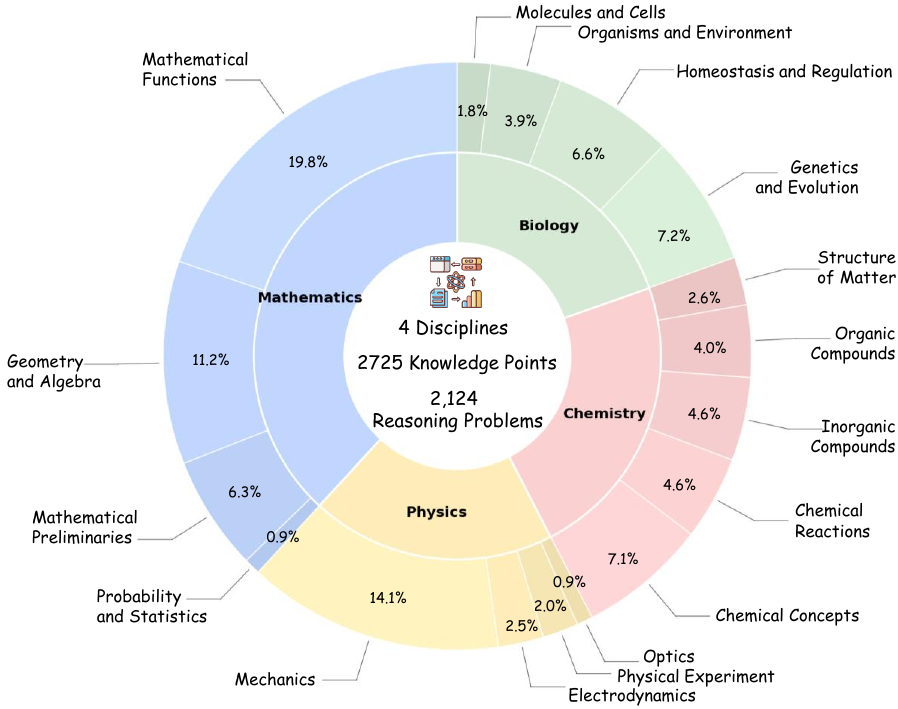}
    \caption{\textbf{Dataset Distribution.} Proportion of questions across disciplines and knowledge points.}
    \label{fig:distribution_chart}
\end{wrapfigure}

According to the criteria and features described above, we incorporate advanced LLMs as pre-annotators to excavate typical questions for these subsets from the whole dataset. Then human experts are prompted to verify these annotations and determine the final composition of 3 subsets. Figure \ref{fig:challenge_subsets} presents illustrative examples from these three challenging subsets, highlighting their distinct inputs and corresponding annotations. Figure \ref{fig:distribution_chart} and Table \ref{tab:dataset_statistics} detail the statistical distribution of the dataset across subjects, modalities, and question types.

\subsection{Dynamic Data Construction Pipeline}

To address the pervasive issue of data contamination and ensure the continuous relevance of our evaluation, we propose a highly automated data construction pipeline based on structured Optical Character Recognition (OCR) and Large Language Model (LLM) parsing. This pipeline systematically processes raw examination PDFs, categorizes and extracts textual and visual elements, and leverages an LLM to decompose the content into structured fields (\textit{e.g.}, question stems, options, standard answers, and reasoning paths) for archiving and subsequent evaluation. Specifically, the dataset construction consists of the following four stages:

\textbf{Examination Paper Collection.} We collected 200 of the latest (published in 2026) authentic Chinese high school examination papers across four disciplines: Mathematics, Physics, Chemistry, and Biology. These freshly curated papers exhibit an extremely low probability of prior data leakage. To establish rigorous ground truths, professional educators meticulously annotated the standard answers and step-by-step reasoning processes. This sourcing strategy ensures fairness across all evaluated 

\noindent models. We specifically selected Chinese examination papers because they are constructed upon a highly scientific and systematic knowledge taxonomy, rigorously tested by tens of millions of students, and widely recognized for their strong representativeness of K-12 educational standards.

\textbf{Structural Document Extraction.} For PDF and scanned image inputs, we utilize a structural extraction workflow to convert the raw exam papers into Markdown text and cropped images. This workflow encompasses preprocessing, text box and embedded figure detection, multimodal OCR (for text, mathematical formulas, and tables), and image cropping. We implement this using the MinerU framework \cite{wang2024mineru}. To better adapt the framework to Chinese examination papers, we fine-tuned the detection and OCR modules on an in-house exam dataset, significantly improving the recognition accuracy of complex mathematical and chemical formulas.

\textbf{LLM-based Parsing with Variable Templates.} Once the raw Markdown text is extracted, it is necessary to identify and structure the individual questions, their corresponding answers, and associated figures. This structured archiving is essential for constructing the context for the evaluated models and for comparing their outputs against the reference answers. Compared to rigid rule-based parsing, utilizing an LLM offers greater flexibility in comprehending diverse paper formats across different subjects and can semantically correct minor OCR recognition errors.

To handle the structural variance and distinct typographical features of exams from different sources, we introduce a variable template-based parsing method. We inject structured descriptions into the LLM's context, including the examination question type, target output fields (format and characteristics), and specific layout features. Before parsing, the template parameters can be adjusted to help the LLM precisely locate and extract the corresponding content. Formally, the context construction and parsing process can be expressed as:
\begin{equation}
    C = \mathcal{T}_{type} \oplus \mathcal{F}_{target} \oplus \mathcal{L}_{layout} \oplus \mathcal{D}_{md}
\end{equation}
\begin{equation}
    S_{db} = \text{LLM}_{\text{parser}}(C)
\end{equation}
where $C$ represents the constructed prompt context, $\mathcal{T}_{type}$ denotes the question type definition, $\mathcal{F}_{target}$ specifies the target JSON schema and field constraints, $\mathcal{L}_{layout}$ describes the source document's layout characteristics, and $\mathcal{D}_{md}$ is the raw extracted Markdown text. The operator $\oplus$ denotes string concatenation, and $S_{db}$ is the final structured JSON database generated by the parsing agent.

\textbf{Verification and De-duplication.} To ensure parsing correctness, the structured text is re-rendered into HTML for manual verification against the original exam papers. This human-in-the-loop review focuses strictly on verifying crucial formulas and problem-solving clues to guarantee that the questions are solvable. Subsequently, we compute text similarity across the problem stems to remove duplicate questions. We extract a hierarchical knowledge tree from the official teaching syllabus and train a knowledge classification model based on Qwen3-VL-4B\cite{Qwen3-VL} to assign knowledge tags to questions without manual annotations.

Ultimately, this extensively automated data ingestion pipeline empowers LiveK12Bench to routinely refresh its test sets and publish refreshed leaderboards, effectively mitigating evaluation biases caused by data contamination. Furthermore, it allows external users to seamlessly upload their own PDF exams to build customized datasets and conduct comprehensive evaluations.

\subsection{Mock-Exam Evaluation Criteria}
\label{sec:mock_exam_evaluation}

Standard reasoning benchmarks typically employ simple accuracy metrics (\textit{e.g.}, Pass@1) to evaluate models. However, this simplified approach fails to authentically reflect an AI's comprehensive performance in human-level examinations. To bridge this gap, we propose a multi-dimensional ``Mock Exam'' evaluation scheme that requires models to complete a full set of test papers under rigorous constraints. Overall, our evaluation criteria encompass four primary dimensions:

\textbf{Outcome Dimension.} For each problem, the evaluated model is prompted to explicitly output its step-by-step reasoning process, followed by the final answer enclosed within a \texttt{\textbackslash boxed\{\}} command. Formally, the generated response $\hat{Y}$ is structured as $\hat{Y} = \mathcal{R} \oplus \text{\textbackslash boxed}\{\hat{A}\}$, where $\mathcal{R}$ denotes the reasoning path, $\hat{A}$ is the extracted final answer. In this dimension, we evaluate the standard Accuracy (Pass@1) metric as $Acc = \sum S_i$, where $S_i = \mathbb{I}(\hat{A}_i \equiv A_i^*)$ for the $i$-th question using an automated evaluator, $A_i^*$ represents the ground-truth answer and $\mathbb{I}(\cdot)$ is the indicator function. Here, the correctness of intermediate steps is ignored. For proof-based questions, the entire generated proof is extracted and compared holistically against the reference.

\textbf{Reasoning Process Dimension.}
Rather than merely inspecting the correctness of the final solutions, we rigorously evaluate the reasoning process to identify potential root-cause errors. Outcomes divorced from their underlying reasoning process have limited evidentiary value. Concretely, the model needs to accurately parse the problem statement, ground its analysis in sound logical assumptions, and conduct progressive deductive reasoning to attain precise solutions. Drawing on a comprehensive analysis of enormous real-world cases, we manually distill three categories of critical reasoning process errors: (1)\textit{Condition Interpretation Error}: a systematic misreading of the given data, relevant theorems, and specified constraints, resulting in a distorted grasp of problem statement facts; (2)\textit{Logical Assumption Error}: the insertion and layering of assumptions or conditions unsubstantiated by the problem statement; (3)\textit{Deductive Reasoning Error}: the applied reasoning steps and computational procedures do not warrant the current inferential claim, thereby manifesting as internal invalidity within the deductive chain and conflicting with established disciplinary laws, theoretical frameworks, or axiomatic systems. We execute a systematic audit of the reasoning process score $P_i$ to detect defined errors, thus proposing a more rigorous standard for evaluation criteria:
\begin{equation}
    P_i = V_i - \tau \cdot \sum_{k=1}^{3} x_{i,k}
\end{equation}
where $V_i$ represents the overall point value for the $i$-th question and $\tau$ denotes the hyperparameter governing the penalty term in the reasoning process evaluation. $x_{i,k} \in \{0,1\}$ indicates the presence of $k$-th reasoning process error.

\textbf{Reasoning Efficiency Dimension.} To quantify the problem-solving efficiency of Large Multimodal Models (LMMs), we introduce two novel metrics:

\noindent \textit{ARL (Accuracy weighted by Response Length):} Inspired by recent studies on efficient reasoning, ARL measures whether a model can achieve high accuracy with a more concise generation. It is formulated as:
\begin{equation}
    \text{ARL} = \frac{1}{N} \sum_{i=1}^{N} S_i \cdot (1 + \lambda\ln\frac{\bar{L}}{l_{i,j}})
    \label{ARL}
\end{equation}
where $N$ is the total number of questions, $\lambda$ denotes the weight factor of reasoning efficiency,$l_{i,j}$ is the generated token length of model $j$ on question $i$, and $\bar{L}$ is an empirical constants of average response length. If a model's generation length aligns with the average level, its ARL equals its Accuracy ($S_i$). An ARL higher than the base Accuracy indicates above-average reasoning efficiency, whereas a lower ARL suggests excessive verbosity.

\noindent \textit{$\text{Acc}_{\leq r}$:} This metric evaluates the model's accuracy when its maximum generation length is strictly restricted to a ratio $r$ of the full context window. By counting the total completion tokens (including intermediate ``thinking'' tokens) as a quantifiable proxy for time, we directly simulate the human capability of finishing an exam within a specific time limit. This approach effectively isolates the impact of varying hardware throughput and model sizes.

\textbf{Exam Performance Dimension.} In this dimension, we meticulously simulate human grading mechanisms to evaluate the authentic test-taking performance of LMMs. The performance is measured by an Overall Exam Score (OES), aggregated from the Exam Score (ES) of individual questions. The ES is composed of a process component ($\text{ES}_i^P$) and an outcome component ($\text{ES}_i^O$):
\begin{equation}
    \text{ES}_i = \underbrace{w_p \cdot P_i\cdot(O_i/V_i)}_{\text{ES}_i^P} + \underbrace{(1 - w_p) \cdot O_i}_{\text{ES}_i^O}
\end{equation}
where $P_i, O_i$ are scores for reasoning and outcome, respectively. $w_p$ balances the weight between process and outcome. Specifically, $O_i$ assigns points equally across correctly answered sub-components; hence, $O_i/V_i$ denotes the proportion of correct sub-questions. We apply this ratio to the score calculation to emphasize the outcome correctness as a prerequisite. Ultimately, the OES is normalized to a standard 100-point scale. We similarly derive the Process Exam Score (PES) and Outcome Exam Score (OCS) by aggregating their respective components:
\begin{equation}
    \text{OES} = 100 \times \frac{\sum \text{ES}_i}{\sum V_i}, \quad \{\text{PES}, \text{OCS}\} = 100 \times \frac{\sum \{\text{ES}_i^P, \text{ES}_i^O\}}{\sum V_i}
    \label{OES}
\end{equation}
By incorporating human-educator-assigned weights and evaluating fine-grained sub-problem correctness, this weighted scoring system distinguishes problem importance and provides a significantly more comprehensive assessment than uniform accuracy.

\begin{table*}[!t]
\centering
\small                    
\setlength{\tabcolsep}{4pt}      
\caption{\textbf{Comparison of LiveK12Bench with existing benchmarks.} }
\label{tab:benchmark_comparison}
\resizebox{\textwidth}{!}{%
\begin{tabular}{@{}lcccccccc@{}}
\toprule
\textbf{Benchmark} & \textbf{Subject(s)} & \textbf{Modality} & \textbf{Kn. Points} & \textbf{Solution} & \textbf{Dynamic} & \textbf{Level} & \textbf{Evaluation} \\ \midrule
MathVista \cite{MathVista} & Math & TI & \textcolor{ForestGreen}{\usym{2713}} & \textcolor{ForestGreen}{\usym{2713}} & \textcolor{red}{\usym{2717}} & K-12,College & A \\
SciBench \cite{wang2023scibench} & Math,Phy,Chem & TO, TI & \textcolor{ForestGreen}{\usym{2713}} & \textcolor{ForestGreen}{\usym{2713}} & \textcolor{red}{\usym{2717}} & College & A \\
M3Exam \cite{zhang2023m3exam} & Multi-subj. & TO, TI & \textcolor{ForestGreen}{\usym{2713}} & \textcolor{red}{\usym{2717}} & \textcolor{red}{\usym{2717}} & K-12 & A \\
GAOKAO-MM \cite{zong2024gaokao} & Multi-subj. & TO, TI & \textcolor{red}{\usym{2717}} & \textcolor{ForestGreen}{\usym{2713}} & \textcolor{red}{\usym{2717}} & High School & A \\
MMSciBench \cite{ye2025mmscibench} & Math, Phy & TO, TI & \textcolor{ForestGreen}{\usym{2713}} & \textcolor{ForestGreen}{\usym{2713}} & \textcolor{red}{\usym{2717}} & High School & A \\
K12Vista \cite{li2025k12vista} & Math,Phy,Chem,Bio & TI & \textcolor{ForestGreen}{\usym{2713}} & \textcolor{ForestGreen}{\usym{2713}} & \textcolor{red}{\usym{2717}} & K-12 & A,P \\
MDK12-Bench \cite{zhou2025mdk12} & Multi-subj. & TO, TI & \textcolor{ForestGreen}{\usym{2713}} & \textcolor{ForestGreen}{\usym{2713}} & \textit{Synthetic} & K-12 & A \\ \midrule
\textbf{LiveK12Bench} & Math,Phy,Chem,Bio & TO,TI,\textbf{IO} & \textcolor{ForestGreen}{\usym{2713}} & \textcolor{ForestGreen}{\usym{2713}} & \textcolor{ForestGreen}{\usym{2713}} & High School & \textbf{A,P,E,Exam} \\ \bottomrule
\end{tabular}%
}
\\[2pt]  
\parbox{\textwidth}{- TO: Text-Only, TI: Text-Image, IO: Image-Only; A: Accuracy, P: Process eval, E: Efficiency eval; Exam: Exam score eval. \textit{Synthetic} denotes achieving dynamic evaluation through synthesizing new questions, in contrast to ours that ingests authentic questions.}
\end{table*}

Finally, to guarantee robust and stable evaluation across all dimensions, we adopt a multi-model arbitration scheme. For all LLM-based evaluators, we aggregate and average the judgments from a panel of multiple advanced models, which substantially mitigates the risk of subjective evaluation bias or comprehension failures from any single judge. Table \ref{tab:benchmark_comparison} provides a comprehensive comparison between LiveK12Bench and existing related benchmarks, underscoring our unique contributions in evaluation dimensions and end-to-end simulation.

\section{Experiments}

In this section, we evaluate a diverse set of Large Multimodal Models (LMMs) on LiveK12Bench. Our experiments are systematically designed to answer four key questions:
(i) How do state-of-the-art models perform across different K-12 disciplines?
(ii) What is the impact of real-world ``snapshot'' noise on reasoning capabilities?
(iii) To what extent are AI models relying on lucky guesses during exams?
(iv) Do LMMs tend to overthink when solving test problems?

\subsection{Experimental Setup}

We test the performance of current mainstream LMMs on the proposed dataset. Our evaluation covers 12 models of varying parameter sizes, categorized into two groups:
\begin{itemize}
    \item \textbf{Proprietaries:} GPT-5\cite{GPT-5}, GPT-5-mini\cite{GPT-5-mini}, Gemini-3-pro\cite{Gemini-3-Pro}, Gemini-3-flash\cite{Gemini-3-Flash}, Claude-opus-4.6\cite{Claude-Opus-4.6}, and Claude-sonnet-4.6\cite{Claude-Sonnet-4.6}, GPT-4o\cite{achiam2023gpt4} (no thinking ability).
    \item \textbf{Open-Source Models:} GLM-5\cite{zeng2026glm}, Kimi-k2.5\cite{team2026kimi}, Qwen3-VL-235B-A22B\cite{Qwen3-VL}, Qwen3-VL-32B\cite{Qwen3-VL}, and Qwen3-VL-8B\cite{Qwen3-VL}.
\end{itemize}

For all problem requests, we employ a unified prompt template, strictly instructing the models to output their final answers in a standardized format (all prompts for LLMs mentioned above are detailed in Appendix \ref{sec:prompts}). For the answer verification process, we utilize a multi-model arbitration panel drawn from four advanced evaluators: GPT-4o, Gemini-3-flash, DeepSeek-V3\cite{deepseek2024}, and Qwen3-30B. These LLMs are selected as evaluators due to their high consistency with human adjustment (over 90\% on average). Specifically, three distinct models are selected for each judgment to prevent any model from evaluating its own generated answers, thereby minimizing self-evaluation bias. Throughout our experiments, the process penalty factor $\tau$ is set as 3, the efficiency weight $\lambda$ and process score weight $\omega_p$ are set as 0.15 and 0.5, respectively. The average-level response length constant $\bar{L}$ is set as 4096 posteriorly according to the statistics of models' responses.
In this section, both scores of the reasoning process (PES) and outcome (OCS) are normalized to a 100-point scale same as OES for better illustration.

\subsection{Results and Analysis}

\textbf{Main results: Disciplinary Performance}

\begin{table*}[!t]
\centering
\small                    
\setlength{\tabcolsep}{4pt}      
\caption{\textbf{Main results on disciplinary performance (26-03 split).}}
\label{tab:main_results}
\resizebox{\textwidth}{!}{%
\begin{tabular}
{@{}l@{\hspace{3pt}}cccccccccccccccc@{}}
\toprule
\multirow{2}{*}{\textbf{Models}}
& \multicolumn{4}{c}{\raisebox{-0.85em}{\includegraphics[height=2.2em]{Figures/subjects/mathematics.png}} \textbf{Mathematics}}
& \multicolumn{4}{c}{\raisebox{-0.7em}{\includegraphics[height=2em]{Figures/subjects/physics.png}} \textbf{Physics}}
& \multicolumn{4}{c}{\raisebox{-0.7em}{\includegraphics[height=2em]{Figures/subjects/chemistry.png}} \textbf{Chemistry}}
& \multicolumn{4}{c}{\raisebox{-0.85em}{\includegraphics[height=2.2em]{Figures/subjects/biology.png}} \textbf{Biology}} \\
\cmidrule(lr){2-5} \cmidrule(lr){6-9} \cmidrule(lr){10-13} \cmidrule(lr){14-17}
& \textbf{Acc} & \textbf{ARL} & \textbf{PES} & \textbf{OES} & \textbf{Acc} & \textbf{ARL} & \textbf{PES} & \textbf{OES} & \textbf{Acc} & \textbf{ARL} & \textbf{PES} & \textbf{OES} & \textbf{Acc} & \textbf{ARL} & \textbf{PES} & \textbf{OES} \\ \midrule
Claude-opus-4.6\textsubscript{\textcolor{teal}{26-02}} &\colorbox{backblue!75}{87.2} &\colorbox{backred!50}{94.6}&87.6 &\colorbox{backblue!75}{90.0} &79.1 &\colorbox{backred!50}{84.2} &80.8 &83.7 &77.0 &\colorbox{backred!50}{81.4} &62.3 &71.1 &\colorbox{backblue!75}{88.0} &\colorbox{backred!50}{91.9} &63.6 &74.1 \\
Gemini-3-pro\textsubscript{\textcolor{teal}{25-11}} &\colorbox{backred!50}{88.3} &82.9&\colorbox{backred!50}{88.3} &\colorbox{backred!50}{90.3} &\colorbox{backred!50}{86.2} &82.5 &\colorbox{backred!50}{85.4} &\colorbox{backred!50}{87.9} &\colorbox{backblue!75}{79.4} &77.2 &\colorbox{backred!50}{71.2} &\colorbox{backred!50}{76.7} &\colorbox{backred!50}{90.7} &89.5 &\colorbox{backred!50}{71.3} &\colorbox{backred!50}{78.6} \\
GPT-5\textsubscript{\textcolor{teal}{25-08}} &82.7 &83.1&83.1 &85.5 &70.9 &70.7 &68.8 &72.7 &59.3 &57.6 &37.9 &44.2 &71.6 &70.3 &44.1 &53.7 \\
Claude-sonnet-4.6\textsubscript{\textcolor{teal}{26-02}} &83.9 &\colorbox{backblue!75}{89.3}&84.5 &87.4 &77.6 &80.4 &78.5 &78.5 &73.8 &75.6 &58.9 &66.4 &84.9 &84.8 &56.1 &65.1 \\
Gemini-3-flash\textsubscript{\textcolor{teal}{25-11}} &\colorbox{backblue!75}{87.2} &84.8&\colorbox{backblue!75}{88.2} &89.8 &\colorbox{backblue!75}{84.4} &82.6 &\colorbox{backblue!75}{85.3} &\colorbox{backblue!75}{87.0} &\colorbox{backred!50}{80.7} &79.7 &\colorbox{backblue!75}{68.6} &\colorbox{backblue!75}{74.4} &\colorbox{backblue!75}{88.0} &86.5 &\colorbox{backblue!75}{68.2} &\colorbox{backblue!75}{75.9} \\
GPT-5-mini\textsubscript{\textcolor{teal}{25-08}} &79.0 & 85.2&75.0 &79.8 &60.1 &63.9 &54.9 &60.1 &45.2 &47.1 &24.1 &29.6 &55.9 &58.5 &31.7 &39.0 \\ \midrule
GLM-5\textsubscript{\textcolor{teal}{26-02}} &79.8 & 76.9&76.5 &79.5 &67.5 &65.3 &56.9 &61.9 &65.1 &63.1 &43.3 &49.7 &77.2 &75.4 &41.1 &52.4 \\
Kimi-k2.5\textsubscript{\textcolor{teal}{26-01}} &85.0 & 86.0&86.2 &87.8 &81.6 &\colorbox{backblue!75}{82.9} &83.3 &85.4 &77.2 &\colorbox{backblue!75}{80.1} &64.0 &71.8 &86.7 &89.7 &64.3 &73.2 \\
Qwen3-VL-235B\textsubscript{\textcolor{teal}{25-09}} &82.5 & 82.5&75.8 &81.3 &77.3 &\colorbox{backblue!75}{82.9} &67.0 &73.8 &74.3 &79.1 &51.6 &62.8 &85.5 &\colorbox{backblue!75}{91.1} &55.0 &65.2 \\
Qwen3-VL-32B\textsubscript{\textcolor{teal}{25-09}} &82.7& 84.3& 76.6 &81.1 & 75.8& 79.2& 69.2 &73.5 &73.8& 77.2& 44.8& 76.2 &86.7& 90.1& 45.9 &55.3 \\
Qwen3-VL-8B\textsubscript{\textcolor{teal}{25-09}} &79.6 &78.8 &70.9 &77.2 &66.6 &65.3 &49.2 &56.8 &62.2&62.0 &30.6& 39.3 &75.3 &74.2 &36.1 &45.5 \\
\addlinespace[2pt]      
\graydashedline         
\addlinespace[2pt]      
GPT-4o\textsubscript{\textcolor{teal}{24-11}} &35.0 &- &19.2 &24.2 &29.1 &- &17.0 &21.0 &29.4 &- &8.0 &11.4 &44.1 &- &16.6 &24.9 \\ \bottomrule
\end{tabular}%
}
\\[2pt]  
\parbox{\textwidth}{- Results of models (grouped into proprietary, open-source, and non-thinking models with \textcolor{teal}{release date}) in 4 subjects using Accuracy (Acc), Efficiency-weighted Accuracy (\hyperref[ARL]{ARL}), Process Exam Score (\hyperref[OES]{PES}), and Overall Exam Score (\hyperref[OES]{OES}). The best and second-best performances are highlighted \textcolor{backred!100}{\textbf{red}} and \textcolor{backblue!100}{\textbf{blue}}.}
\end{table*}

Table \ref{tab:main_results} presents the performance of the evaluated models across different disciplines on the full dataset, assessing LMMs' capabilities in realistic exam distributions. Gemini-3-pro achieves the highest Accuracy (Acc) and Overall Exam Score (OES) across most subjects. Its notable advantage is particularly evident in the Process Exam Score (PES) for Chemistry and Biology, outperforming the second-best model by 2.6 and 3.1 points, respectively. Notably, the smaller Gemini-3-flash also attains highly competitive scores. When considering reasoning efficiency, Claude-opus-4.6 achieves the highest ARL scores across all disciplines, indicating it attains higher accuracy with fewer reasoning tokens. Conversely, Gemini models yield relatively lower ARL scores (ranking 5th to 7th), reflecting a design trade-off that sacrifices reasoning time for enhanced performance. Furthermore, while open-source models like Kimi-k2.5 and Qwen3 exhibit a slight disadvantage in exam scores compared to leading proprietary models, their reasoning efficiency surpasses that of GPT-5 and Gemini. It is also observable that factors such as smaller parameter sizes, earlier release dates, or a lack of explicit ``thinking'' capabilities cause models like GPT-4o (which is excluded from ARL evaluation due to the absence of thinking tokens) and Qwen3-VL-8B to struggle significantly at lower performance tiers. Comparing across subjects, all models perform worse in Chemistry and Biology compared to Mathematics and Physics---the traditional focus of mainstream reasoning research. This performance drop stems primarily from broader knowledge taxonomies and more complex Q\&A structures involving multiple images and sub-questions.

\begin{table}[!t]
\centering
\caption{\textbf{Performance on challenging subsets.}}
\label{tab:combined}
\begin{minipage}[t]{0.49\textwidth}
\centering
{(a) Complex Layout Set} \\[7pt]
\small
\setlength{\tabcolsep}{3pt}
\resizebox{\linewidth}{!}{%
\begin{tabular}{@{}lcccc@{}}
\toprule
\multirow{2}{*}{\textbf{Models}} & \multicolumn{2}{c}{\textbf{Acc}} & \multicolumn{2}{c}{\textbf{OCS}} \\
\cmidrule(lr){2-3} \cmidrule(lr){4-5}
 & Standard & Exam & Standard & Exam \\ \midrule
Claude-opus-4.6 & 87.2 & 55.0\textsubscript{\textcolor{red}{-32.2}} & 92.8 & 50.6\textsubscript{\textcolor{red}{-42.2}} \\
Gemini-3-pro & 88.3 & \colorbox{backred!50}{60.0}\textsubscript{\textcolor{red}{-28.3}} & 92.6 & \colorbox{backred!50}{53.7}\textsubscript{\textcolor{red}{-38.9}} \\
GPT-5 & 82.7 & 50.4\textsubscript{\textcolor{red}{-32.3}} & 88.1 & 44.1\textsubscript{\textcolor{red}{-44.0}} \\
Claude-sonnet-4.6 & 83.9 & 45.5\textsubscript{\textcolor{red}{-38.4}} & 90.8 & 45.8\textsubscript{\textcolor{red}{-45.0}} \\
Gemini-3-flash & 87.2 & 55.0\textsubscript{\textcolor{red}{-32.2}} & 91.8 & 50.0\textsubscript{\textcolor{red}{-41.8}} \\
GPT-5-mini & 79.0 & 42.9\textsubscript{\textcolor{red}{-36.1}} & 85.7 & 39.8\textsubscript{\textcolor{red}{-45.9}} \\ \midrule
GLM-5 & 79.8 & - & 85.4 & - \\
Kimi-k2.5 & 85.0 & \colorbox{backblue!75}{59.0}\textsubscript{\textcolor{red}{-26.0}} & 90.2 & \colorbox{backblue!75}{51.8}\textsubscript{\textcolor{red}{-38.4}} \\
Qwen3-VL-235B & 82.5 & 56.0\textsubscript{\textcolor{red}{-26.5}} & 88.4 & 50.6\textsubscript{\textcolor{red}{-37.8}} \\
Qwen3-VL-32B & 82.7 & 55.5\textsubscript{\textcolor{red}{-27.2}} & 85.6 & 48.2\textsubscript{\textcolor{red}{-37.4}} \\
Qwen3-VL-8B & 79.6 & 53.6\textsubscript{\textcolor{red}{-26.0}} & 83.5 & 46.8\textsubscript{\textcolor{red}{-36.7}} \\
\addlinespace[2pt]      
\graydashedline         
\addlinespace[2pt]      
GPT-4o & 35.0 & 5.5\textsubscript{\textcolor{red}{-29.5}} & 40.8 & 6.6\textsubscript{\textcolor{red}{-34.2}} \\ \bottomrule
\end{tabular}%
}
\end{minipage}%
\hfill
\begin{minipage}[t]{0.49\textwidth}
\centering
{(b) Rigorous Process Set} \\[2pt]
\small                    
\setlength{\tabcolsep}{4pt}      
\resizebox{\linewidth}{!}{
\begin{tabular}{@{}lcccccc@{}}  
\toprule
\multirow{2}{*}{\textbf{Models}} & \multicolumn{3}{c}{Exam Score $\uparrow$} & \multicolumn{3}{c}{Process Error $\downarrow$} \\
\cmidrule(lr){2-4} \cmidrule(lr){5-7}
& \textbf{OCS} & \textbf{PES} & \textbf{OES} & \textbf{CIE} & \textbf{LAE} & \textbf{DRE} \\ \midrule
Claude-opus-4.6 & 89.0 & 68.7 & 78.0 & 36 & \colorbox{backblue!75}{7} & \colorbox{backblue!75}{11} \\
Gemini-3-pro & \colorbox{backblue!75}{90.9} & \colorbox{backred!50}{76.9} & \colorbox{backred!50}{81.7} & \colorbox{backred!50}{17} & \colorbox{backblue!75}{7} & \colorbox{backred!50}{10} \\
GPT-5 & 83.0 & 56.0 & 61.0 & 61 & 13 & 30 \\
Claude-sonnet-4.6 & 87.9 & 63.9 & 70.8 & 40 & \colorbox{backred!50}{6} & 20 \\
Gemini-3-flash & \colorbox{backred!50}{91.6} & \colorbox{backblue!75}{75.4} & \colorbox{backblue!75}{80.3} & \colorbox{backred!50}{17} & 8 & 18 \\
GPT-5-mini & 75.8 & 42.7 & 48.7 & 78 & 29 & 46 \\ \midrule
GLM-5 & 77.1 & 54.1 & 60.4 & 46 & 31 & 23 \\
Kimi-k2.5 & 87.5 & 71.9 & 77.3 & 25 & \colorbox{backblue!75}{7} & 16 \\
Qwen3-VL-235B & 87.0 & 63.6 & 70.3 & 36 & 12 & 25 \\
Qwen3-VL-32B & 87.6 & 52.4 & 59.7 & 71 & 18 & 59 \\
Qwen3-VL-8B & 79.7 & 44.2 & 50.9 & 47 & 11 & 36 \\
\addlinespace[2pt]      
\graydashedline         
\addlinespace[2pt]      
GPT-4o & 50.0 & 13.5 & 19.8 & 129 & 34 & 111 \\ \bottomrule
\end{tabular}%
}
\end{minipage}
\\[5pt]  
\parbox{\textwidth}{- (a) \hyperref[OES]{OCS}: Outcome exam score, Standard modality: text-only or text-image input, Exam modality: Image-only input. The score differences between two modalities are noted in \textcolor{red}{red}. (b) PES: Process Exam Score, OES: Overall Exam Score. We report the number of process errors in 3 types (CIE: condition interpretation error, LAE: logical assumption error, DRE: deductive reasoning error).}
\end{table}

\textbf{What is the impact of real-world ``snapshot'' noise on reasoning capabilities?}
Table \ref{tab:combined} (Left) details the model performance on the Complex Layout Set. We contrast performance under standard input modalities (parsed TO/TI) against the Exam modality (IO) to investigate how LMMs handle the complex visual layouts of raw exam papers. Significantly, unparsed layouts and embedded images cause a drastic decline in both Acc and Outcome Score (OCS) across all models. The GPT series is the most severely impacted, with an average drop of 33.7\% in Acc and 44.9 points in OCS. In contrast, Gemini and Kimi-k2.5 demonstrate superior visual robustness. According to the LLM-judge analysis, errors predominantly originate from image dislocation and complex text-image interleaving, which lead models to overlook or misinterpret crucial visual information, thereby derailing subsequent reasoning steps.

\textbf{To what extent do AIs rely on lucky guesses during exams?}
Table \ref{tab:combined} (Right) evaluates the Rigorous Process Set. Beyond exam scores, we tally the frequencies of the three reasoning process errors defined in Section \ref{sec:mock_exam_evaluation}. Compared to the general distribution in Table \ref{tab:main_results}, all models exhibit lower PES but maintain relatively high OCS. This divergence quantitatively demonstrates that for these specific problems, models frequently arrive at the correct final answer through flawed reasoning---effectively relying on ``lucky guesses.'' Comparing the three error types, models are more prone to Condition Interpretation Errors (CIE) and Deductive Reasoning Errors (DRE), while Logical Assumption Errors (LAE, or assumption hallucinations) are less frequent. Notably, the Gemini series maintains higher process quality, which fundamentally accounts for its leading overall scores in Table \ref{tab:main_results}.

\begin{figure*}[!t]
    \centering
    \includegraphics[width=\textwidth]{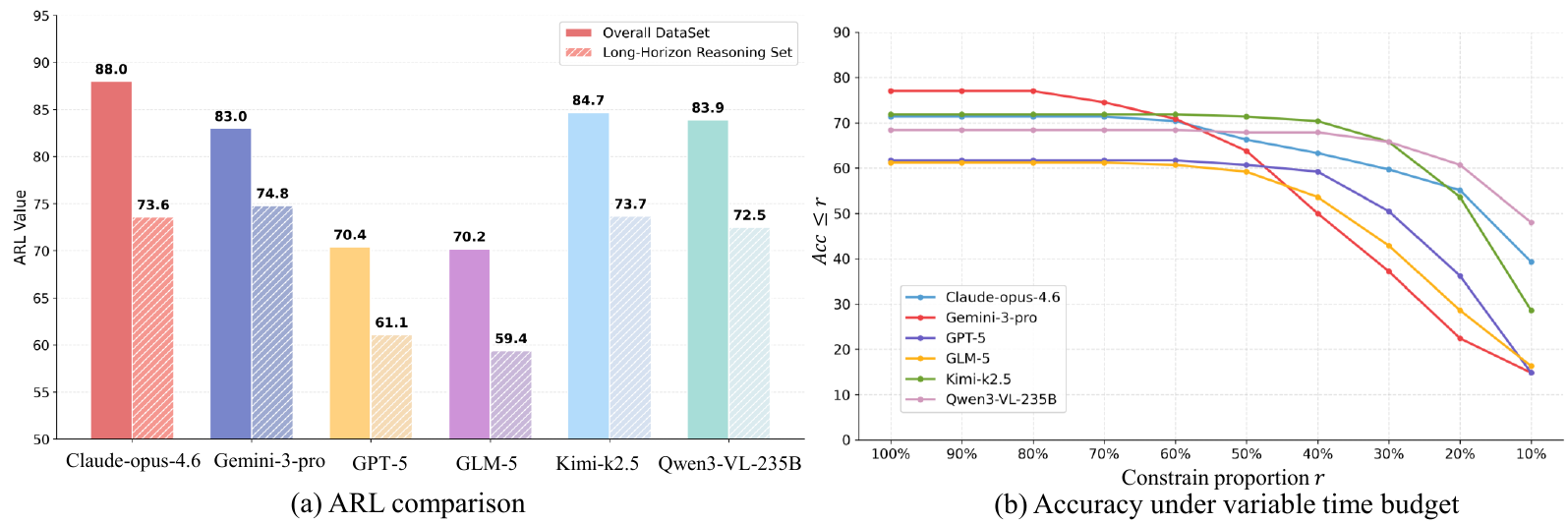}
    \caption{\textbf{Reseults comparison on Long-horizon Reasoning Set.}}
    \label{fig:long_horizon}
\end{figure*}

\textbf{Do LMMs overthink during exams?}
Figure \ref{fig:long_horizon}(a) compares the ARL performance of LMMs on the long-horizon reasoning subset with that on the overall dataset. We observe significant ARL score degradation across all models (ranging from -8 to -15), highlighting the reasoning complexity of this subset constructed via the strategy outlined in Section \ref{long_reason}. Notably, Gemini-3-pro exhibits the smallest performance drop (-8.2) and achieves the highest score (74.8) on this subset, despite ranking only fourth on the overall dataset. This suggests that Gemini-3-pro is less prone to overthinking than Claude-Opus-4.6 when facing the peak complexity in reasoning tasks, achieving higher accuracy through a more efficient allocation of its reasoning budget.

Figure \ref{fig:long_horizon}(b) illustrates the accuracy variation of models on the Long-Horizon Reasoning Set as the time limit (proportional threshold of context) decreases. This aims to assess the usability of LMMs under varying speed requirements. As the generation budget is restricted to 10\% of the default length (i.e., 3.2k tokens), most models experience significant timeout failures, causing their accuracy to plummet to approximately half of their unconstrained performance. Qwen3-VL exhibits the smallest accuracy drop, indicating it can maintain robust capabilities even under stringent token budgets, making it highly suitable for scenarios requiring rapid responses. Conversely, Gemini-3-pro experiences the earliest and most precipitous decline in accuracy. This suggests that while it excels when allowed to deliberate extensively, its practical utility diminishes in fast-response applications. Detailed problem examples and model responses for all subsets are provided in Appendix \ref{sec:examples}.

\section{Related Works}

\subsection{Scientific Reasoning Benchmarks for LMMs}

The evaluation of Large Multimodal Models (LMMs) on scientific problems initially expanded from the discipline of mathematics to encompass broader scientific domains, including physics, chemistry, informatics and general science problems \cite{jassim2023grasp,li2024mmscibench,wang2023scibench,evans2023msvec,tarsi2024sciol}.
In particular, research focusing on high-school and pre-college levels has progressively evolved into capability assessments tailored for K-12 educational scenarios\cite{zhang2023evaluating,zong2024gaokao,he2024olympiadbench}. By benchmarking models against problems designed for human students, these studies serve as a natural testbed for their potential applications in the educational sector, such as intelligent tutoring systems and AI assistants \cite{kasneci2023chatgpt}.

While some existing benchmarks, such as K12Vista \cite{li2025k12vista}, MDK12Bench \cite{zhou2025mdk12}, MMSciBench \cite{ye2025mmscibench}, have proposed evaluation methodologies for K-12 level problems, they typically evaluate questions in isolation. Other exam-oriented benchmarks like M3Exam \cite{zhang2023m3exam} and Exams-v \cite{das2024exams} focus on multi-lingual evaluation.
Human education, however, is heavily anchored in comprehensive examination systems that encapsulate massive educational resources and standardized curricula. Existing works lack a systematic evaluation of model capabilities within these authentic K-12 examination scenarios. In contrast, LiveK12Bench addresses this gap by providing a holistic and systematic examination-based evaluation protocol for LMMs.


The evaluation of reasoning capabilities has consistently been a pivotal driving force behind the rapid development of large reasoning models. From early text-based benchmarks like GSM8K \cite{cobbe2021training} and MATH \cite{hendrycks2021measuring} to recent multimodal benchmarks evaluating visual reasoning such as MathVista, MathVision, etc. \cite{lu2024mathvista,Math-Vision,MathVerse,We-Math}, a crucial dimension among these is assessing whether a model can successfully extract valid and accurate information from visual inputs to construct the premises for subsequent logical deductions.
Driven by real-world practical scenarios, we introduce a novel visual reasoning challenge: parsing problem information directly from the visually noisy layouts of raw examination pages, accurately comprehending image properties, and formulating the correct premises for reasoning.

\section{Conclusions and Future Directions}

In this paper, we introduce LiveK12Bench, a dynamic, comprehensive, and multi-disciplinary benchmark designed to authentically simulate human K-12 examinations. Powered by a highly efficient and automated data ingestion pipeline, our framework continuously exposes state-of-the-art LMMs to the recent and uncontaminated real-world exam questions. By restricting models to an end-to-end ``Image-Only'' input modality under time constraints, we systematically evaluate their abilities regarding visual robustness, process rigor, and reasoning efficiency. Utilizing multi-dimensional criteria modeled after professional educators' grading standards, we assign a holistic overall exam score to each evaluated model. Our evaluation reveals that even the most advanced models, including the GPT-5 series, exhibit substantial space for improvement when confronted with complex visual layouts and rigorous reasoning process assessments.

Based on our findings, future research should focus on two areas. First, models need to better handle complex visual scenes and provide helpful, heuristic solutions for learners, not just correct answers. Second, evaluations should test pedagogical effectiveness and handle real-world inputs like handwritten photos. Moving forward, LiveK12Bench will continuously expand its scale by periodically integrating the latest, uncontaminated exam papers through our automated pipeline. The complete dataset and evaluation codebase will be open-sourced upon publication to foster further advancements in educational AI.

\newpage
\bibliography{main_arxiv}
\newpage

\providecommand{\graydashedline}{\arrayrulecolor{gray!80}\hdashline\arrayrulecolor{black}}

\newtcolorbox{promptbox}[1]{%
  enhanced,
  breakable,
  colback=gray!5,
  colframe=gray!50,
  colbacktitle=gray!55,
  coltitle=white,
  fonttitle=\bfseries\small,
  title={#1},
  left=6pt, right=6pt, top=4pt, bottom=4pt,
  boxrule=0.5pt,
}

\providecommand{\ForestGreen}{}
\definecolor{ForestGreen}{RGB}{0,235,110}
\definecolor{backred}{RGB}{255, 190, 190}
\definecolor{backblue}{RGB}{220, 230, 250}
\definecolor{myred}{HTML}{FFB3B3}
\newcommand{\hlred}[1]{\sethlcolor{myred}\hl{#1}}

\appendix

\section{Supplementary Results}

Table \ref{tab:answer_hack},\ref{tab:long_reason},\ref{tab:complex_layout} supplement more complete results of LMMs on three challenging subsets. Overall, the relative performance gaps between different LMMs remain the same as shown in Table 3 in the body part, where Gemini-3-pro and Claude-opus-4.6 rank at the forefront and Kimi-k2.5 shows advantages among open-source models. From the perspective of subsets, the complete results consistently verify the challenges of the three subsets: easy to overlook important processes, require long-term reasoning, and are prone to misperceiving visual conditions (see high CIE score in Table \ref{tab:complex_layout}).

\begin{table*}[!h]
\centering
\small
\setlength{\tabcolsep}{8pt}
\caption{\textbf{Results on the Rigorous Process Set.}}
\label{tab:answer_hack}
\resizebox{\textwidth}{!}{%
\begin{tabular}{@{}lccccccccc@{}}
\toprule
\textbf{Models} & \textbf{Acc} & \textbf{ARL} & \textbf{$\text{Acc}_{\leq 50}$} & \textbf{PES} & \textbf{OCS} & \textbf{OES} & \textbf{CIE$\downarrow$} & \textbf{LAE$\downarrow$} & \textbf{DRE$\downarrow$} \\ \midrule
Claude-opus-4.6 & 85.1 & \colorbox{backred!50}{89.8} & 82.9 & 68.7 & 89.0 & 78.0 & 32 & \colorbox{backblue!75}{6} & \colorbox{backblue!75}{10} \\
Gemini-3-pro & \colorbox{backred!50}{88.7} & 87.2 & \colorbox{backred!50}{87.4} & \colorbox{backred!50}{76.9} & \colorbox{backblue!75}{90.9} & \colorbox{backred!50}{81.7} & \colorbox{backred!50}{15} & \colorbox{backblue!75}{6} & \colorbox{backred!50}{9} \\
GPT-5 & 73.9 & 72.7 & 73.4 & 56.0 & 83.0 & 61.0 & 55 & 12 & 27 \\
Claude-sonnet-4.6 & 81.1 & 82.9 & 80.2 & 63.9 & 87.9 & 70.8 & 36 & \colorbox{backred!50}{5} & 18 \\
Gemini-3-flash & \colorbox{backblue!75}{87.8} & 85.1 & 82.4 & \colorbox{backblue!75}{75.4} & \colorbox{backred!50}{91.6} & \colorbox{backblue!75}{80.3} & \colorbox{backred!50}{15} & 7 & 16 \\
GPT-5-mini & 63.1 & 66.4 & 63.1 & 42.7 & 75.8 & 48.7 & 70 & 26 & 41 \\
\midrule
GLM-5 & 71.6 & 69.4 & 70.3 & 54.1 & 77.1 & 60.4 & 41 & 28 & 21 \\
Kimi-k2.5 & 85.1 & 87.9 & \colorbox{backblue!75}{85.1} & 71.9 & 87.5 & 77.3 & \colorbox{backblue!75}{23} & \colorbox{backblue!75}{6} & 14 \\
Qwen3-VL-235B & 82.9 & 88.1 & 82.4 & 63.6 & 87.0 & 70.3 & 32 & 11 & 23 \\
Qwen3-VL-32B & 85.6 & \colorbox{backblue!75}{89.1} & \colorbox{backblue!75}{85.1} & 52.4 & 87.6 & 59.7 & 42 & 10 & 32 \\
Qwen3-VL-8B & 72.5 & 71.8 & 71.6 & 44.2 & 79.7 & 50.9 & 64 & 18 & 53 \\
\addlinespace[2pt]
\graydashedline
\addlinespace[2pt]
GPT-4o & 36.5 & 36.5 & 36.5 & 13.5 & 50.0 & 19.8 & 116 & 31 & 100 \\
\bottomrule
\end{tabular}%
}
\\[2pt]
\parbox{\textwidth}{- The best and second-best performances are highlighted \textcolor{backred!100}{\textbf{red}} and \textcolor{backblue!100}{\textbf{blue}}. $\downarrow$ indicates lower is better (error counts).}
\end{table*}

\begin{table*}[!h]
\centering
\small
\setlength{\tabcolsep}{8pt}
\caption{\textbf{Results on the Long-horizon Reasoning Set.}}
\label{tab:long_reason}
\resizebox{\textwidth}{!}{%
\begin{tabular}{@{}lccccccccc@{}}
\toprule
\textbf{Models} & \textbf{Acc} & \textbf{ARL} & \textbf{$\text{Acc}_{\leq 50}$} & \textbf{PES} & \textbf{OCS} & \textbf{OES} & \textbf{CIE$\downarrow$} & \textbf{LAE$\downarrow$} & \textbf{DRE$\downarrow$} \\ \midrule
Claude-opus-4.6 & 71.4 & \colorbox{backblue!75}{74.8} & 66.3 & 67.7 & 83.3 & 73.8 & 32 & \colorbox{backblue!75}{7} & \colorbox{backblue!75}{9} \\
Gemini-3-pro & \colorbox{backblue!75}{74.5} & 73.2 & \colorbox{backred!50}{72.4} & \colorbox{backblue!75}{71.4} & \colorbox{backblue!75}{83.4} & \colorbox{backblue!75}{76.6} & \colorbox{backred!50}{17} & \colorbox{backblue!75}{7} & \colorbox{backblue!75}{9} \\
GPT-5 & 61.7 & 61.1 & 60.7 & 55.5 & 77.8 & 61.3 & 48 & 11 & 36 \\
Claude-sonnet-4.6 & 73.0 & \colorbox{backred!50}{75.4} & 70.4 & 67.5 & 83.0 & 72.1 & 36 & \colorbox{backred!50}{5} & 17 \\
Gemini-3-flash & \colorbox{backred!50}{77.0} & 73.6 & 63.8 & \colorbox{backred!50}{74.1} & \colorbox{backred!50}{86.6} & \colorbox{backred!50}{78.6} & \colorbox{backred!50}{17} & 10 & \colorbox{backred!50}{6} \\
GPT-5-mini & 47.4 & 50.9 & 47.4 & 40.5 & 68.2 & 46.3 & 77 & 28 & 53 \\
\midrule
GLM-5 & 61.2 & 59.4 & 59.2 & 48.3 & 73.1 & 55.0 & 45 & 33 & 30 \\
Kimi-k2.5 & 71.9 & 73.7 & \colorbox{backblue!75}{71.4} & 69.3 & 83.0 & 75.3 & \colorbox{backblue!75}{19} & 10 & 19 \\
Qwen3-VL-235B & 68.4 & 72.5 & 67.9 & 51.0 & 79.7 & 60.9 & 41 & 19 & 37 \\
Qwen3-VL-32B & 67.0 & 71.5 & 66.5 & 41.0 & 80.5 & 51.5 & 54 & 17 & 51 \\
Qwen3-VL-8B & 54.5 & 54.0 & 54.0 & 34.5 & 72.0 & 42.0 & 83 & 29 & 80 \\
\addlinespace[2pt]
\graydashedline
\addlinespace[2pt]
GPT-4o & 21.4 & 21.4 & 21.4 & 8.8 & 37.7 & 13.3 & 107 & 51 & 128 \\
\bottomrule
\end{tabular}%
}
\end{table*}

\begin{table*}[!h]
\centering
\small
\setlength{\tabcolsep}{8pt}
\caption{\textbf{Results on the Complex Layout Set in Image-Only Modality Setting.}}
\label{tab:complex_layout}
\resizebox{\textwidth}{!}{%
\begin{tabular}{@{}lcccccccc@{}}
\toprule
\textbf{Models} & \textbf{Acc} & \textbf{ARL} & \textbf{PES} & \textbf{OCS} & \textbf{OES} & \textbf{CIE$\downarrow$} & \textbf{LAE$\downarrow$} & \textbf{DRE$\downarrow$} \\ \midrule
Claude-opus-4.6 & 55 & \colorbox{backred!50}{57.1} & \colorbox{backred!50}{44.9} & 50.6 & \colorbox{backred!50}{47.7} & \colorbox{backblue!75}{43} & \colorbox{backblue!75}{11} & \colorbox{backblue!75}{8} \\
Gemini-3-pro & \colorbox{backred!50}{60} & 56.5 & 44.2 & \colorbox{backred!50}{53.7} & 46.8 & 44 & 10 & 9 \\
GPT-5 & 38 & 37.5 & 27 & 44.1 & 29.9 & 53 & 17 & 21 \\
Claude-sonnet-4.6 & 45.5 & 46.2 & 33.5 & 45.8 & 36.5 & 48 & 10 & 15 \\
Gemini-3-flash & 55 & \colorbox{backblue!75}{52.8} & \colorbox{backblue!75}{40.3} & 50 & \colorbox{backblue!75}{44.1} & \colorbox{backred!50}{42} & \colorbox{backred!50}{9} & \colorbox{backred!50}{7} \\
GPT-5-mini & 30.5 & 30.2 & 18.5 & 39.8 & 20.8 & 68 & 35 & 33 \\
\midrule
Kimi-k2.5 & \colorbox{backblue!75}{59} & 57.0 & 44.0 & \colorbox{backblue!75}{51.8} & 46.5 & 31 & 10 & 11 \\
Qwen3-VL-235B & 56 & 56.2 & 40.0 & 50.6 & 44.0 & 44 & 18 & 18 \\
Qwen3-VL-32B & 55.5 & 55.8 & 33.0 & 48.2 & 37.5 & 56 & 17 & 26 \\
Qwen3-VL-8B & 53.6 & 36.5 & 22.5 & 46.8 & 26.0 & 63 & 27 & 43 \\
\addlinespace[2pt]
\graydashedline
\addlinespace[2pt]
GPT-4o & 5.5 & 5.5 & 2.0 & 6.6 & 3.0 & 118 & 43 & 82 \\
\bottomrule
\end{tabular}%
}
\end{table*}

\section{Detailed Prompts}
\label{sec:prompts}

\begin{promptbox}{Problem-Solving System Prompt}
\small
Please solve the following problem. Present your step-by-step solution process and mark your final answer according to the following rules:\par\smallskip
(1) For single-choice questions, place the correct option letter inside \verb|\boxed{}|, e.g., $\boxed{\text{A}}$;\par
(2) For multiple-choice questions, place all correct option letters inside a single \verb|\boxed{}|, e.g., $\boxed{\text{ACD}}$;\par
(3) For fill-in-the-blank questions, enclose each blank's answer in a separate \verb|\boxed{}|, e.g., for one blank: $\boxed{x}$; for two blanks: $\boxed{x}$ and $\boxed{y}$;\par
(4) For open-ended questions, enclose each sub-question's answer in a separate \verb|\boxed{}|, e.g., for one sub-question: $\boxed{x}$; for two sub-questions: $\boxed{x}$ and $\boxed{y}$.
\end{promptbox}

\begin{promptbox}{Answer Evaluation Prompt}
\small
You are an expert grader. Determine whether the student's answer is consistent with the standard answer, i.e., whether the student answered correctly. Below are the grading criteria:\par\smallskip
1. Question types include fill-in-the-blank and open-ended questions. Some questions may contain multiple sub-questions, such as fill-in-the-blank questions with multiple blanks or open-ended questions with multiple parts. You must judge each sub-question independently.\par
2. Answers may be expressed in different forms---for example, as a mathematical expression or a textual description. As long as the semantic meaning is equivalent, the answer is considered correct. Equivalent formulas expressed in different notations are also accepted. If equivalence cannot be determined, mark the student's answer as incorrect.\par
3. You do not need to re-derive the answer, as the standard answer is already provided. Simply compare the student's answer with the standard answer based on the question format to determine correctness.\par
4. For questions with definitive results, compare the student's final answer (typically enclosed in \verb|\boxed{}|) with the standard answer; the solution process need not be evaluated. For questions without definitive results (e.g., proofs), evaluate whether the solution approach is correct (a sound reasoning approach suffices).\par\smallskip
Based on the above criteria, output the number of sub-questions the student answered correctly, enclosed in \verb|\boxed{}|. For example: if there is only one question and the answer is correct, output $\boxed{1}$; if there are 3 sub-questions and the student answered 2 correctly, output $\boxed{2}$.\par\smallskip
Below is the content for your evaluation:\par
\textbf{Original Question:} \texttt{\{question\}}\par
\textbf{Standard Answer and Solution:} \texttt{\{gold\_answer\}} \texttt{\{gold\_solution\}}\par
\textbf{Student's Answer:} \texttt{\{answer\}}\par\smallskip
\end{promptbox}

\begin{promptbox}{Process Evaluation Prompt}
\small
As an expert evaluator of science problem solutions, you will receive a test question, a standard solution process, and a student's solution process, with the following specific content:\par\smallskip
\textbf{Question:} \texttt{\{question\}}\par
\textbf{Standard Solution:} \texttt{\{gold\_solution\}}\par
\textbf{Student's Solution:} \texttt{\{answer\}}\par\smallskip
Your task is to examine the student's solution sentence by sentence, identify erroneous steps by comparison with the standard solution, and output the error count and error types.\par\smallskip
Only three types of objective errors are considered:\par\smallskip
(1) \textbf{Condition Interpretation Error (CIE):} An inconsistency in numerical information or domain knowledge relative to the information embedded in the question's figures, fundamental axioms/theorems of the relevant domain, or the basic conditions stated in the problem. This error leads to a systematic misunderstanding of the original problem's factual statements, variable semantics, conditional relationships, or solution objectives.\par\smallskip
(2) \textbf{Logical Assumption Error (LAE):} Failure to fully utilize the constraints and data given in or directly derivable from the problem statement, while introducing unsupported additional assumptions, simplifications, or default conditions, and reasoning based upon them, causing systematic deviation in reasoning direction or conclusions. The essence of this error lies in the combination of insufficient information utilization and unfounded assumptions.\par\smallskip
(3) \textbf{Deductive Reasoning Error (DRE):} During the derivation from existing conditions, the reasoning steps, computational operations, or logical structures employed fail to support the current step's conclusion. This error manifests as incorrectness or imprecision within the deductive chain, conflicting with established disciplinary laws, theoretical frameworks, or axiomatic systems, resulting in conclusions that are inconsistent with existing knowledge.\par\smallskip
Please compare the student's solution process with the standard solution according to the above error definitions, and strictly identify the corresponding erroneous steps. A single question may simultaneously contain multiple types of errors.\par\smallskip
\textbf{Output format:} First output the total number of source error steps, then separately output whether each of the three error types exists (1 = present, 0 = absent) for (1) CIE, (2) LAE, (3) DRE, each enclosed in \verb|\boxed{}|. If there are no errors, output $\boxed{0}\ \boxed{0}\ \boxed{0}\ \boxed{0}$.\par\smallskip
\textbf{Note:} Only count source errors; errors propagated from previous erroneous steps should not be counted.\par\smallskip
\end{promptbox}

\begin{promptbox}{Exam Paper Parsing Prompt}
\small
You are a meticulous and error-free document organization assistant. Based on the provided document text, extract the required information and output it in JSON format.\par\smallskip
The document contains multiple questions. Organize each item's information into a \texttt{list(dict)} structure. Each \texttt{dict} must contain the following keys with their respective requirements: \texttt{\{target\_items\}}. Return \texttt{null} for any key whose information cannot be found.\par\smallskip
To facilitate locating the target information, the input is likely to follow this format: \texttt{\{paper\_template\}}. If the document does not conform to the above format, analyze its content flexibly and complete the task.\par\smallskip
\textbf{Notes:}\par
1. Do not modify any document content (including text content and LaTeX formula content); only extract and organize the corresponding information.\par
2. Ensure all questions in the document are parsed without omission.\par
3. Output only a result that can be directly parsed as valid JSON without errors.\par
4. If there is no more content to extract, output only \texttt{[DONE]}.\par\smallskip
\textbf{Target Schema} (\texttt{target\_items}):\par\smallskip
\texttt{question\_types} = [\texttt{"multiple-choice"}, \texttt{"fill-in-the-blank"}, \texttt{"open-ended"}]\par\smallskip
\begin{itemize}\setlength{\itemsep}{2pt}
\item \texttt{type}: A string value. Select one from \texttt{question\_types} based on the section header and question content. Multiple-choice questions typically contain options A/B/C/D; fill-in-the-blank questions feature underlined blanks; open-ended questions pose problems with possible sub-questions.
\item \texttt{score}: An integer value. Determined by the prefix of each question or the section header. Multiple-choice and fill-in-the-blank are typically $\leq$ 5 points; open-ended are typically 5--20 points.
\item \texttt{question}: A string value. The question text beginning with a numeric index, conforming to its question type pattern. \textbf{Must not include answer information or image paths.}
\item \texttt{answer}: A \texttt{list(str)} structure. Each sub-question's answer corresponds to one string element. For multiple-select answers, all selected options are placed in a single string, e.g., answer \texttt{"ABC"} should be extracted as \texttt{["ABC"]}.
\item \texttt{solution}: A string value. The solution process, usually identified by a distinctive marker or field header.
\item \texttt{images}: A \texttt{list(str)} of image storage paths associated with the question, identified by the pattern: \texttt{![](<image\_path>)}. \textbf{Do not fabricate paths.}
\end{itemize}
\smallskip
\textbf{Paper Template} (\texttt{paper\_template}):\par\smallskip
The document is an exam paper with solutions. The target objects are individual multiple-choice and open-ended questions that contain the target fields (ignore questions without solutions and other content).\par\smallskip
For each question: the question type is indicated in the section header with Chinese numerals; the score appears in parentheses before the question or in the section header; questions typically begin with Arabic numerals (1, 2, 3, etc.); answers are found within the solution section.
\end{promptbox}

\section{Reliability and Stability of LLM-as-Judge Process Evaluation}
\label{sec:appendix-judge-reliability}

Process evaluation in LiveK12Bench relies on a multi-LLM arbitration panel rather than a single judge. To validate that this design produces trustworthy Process Exam Scores (PES), we perform two complementary studies: (i) a direct comparison against human-expert judgments, and (ii) a stability analysis over repeated evaluations, alternative judge panels, and dataset partitions.

\subsection{Accuracy Relative to Human Expert Judgments}
\label{sec:appendix-judge-human}

We recruited two PhD students with strong K-12 STEM backgrounds to independently identify process errors on a stratified sample drawn from the Rigorous Process Set. Their annotations form the human reference. Table~\ref{tab:pes-human} reports per-error-type accuracy, precision, recall, F1, and Cohen's $\kappa$ between each evaluator setup and the human reference, alongside the inter-annotator agreement between the two human experts as an empirical upper bound.

\begin{table}[!htbp]
\centering
\caption{\textbf{Accuracy of LLM judges relative to human-expert process-error annotations.} Multi-LLM arbitration approaches the human-human upper bound on every metric while substantially outperforming a single-judge baseline.}
\label{tab:pes-human}
\small                                   
\renewcommand{\arraystretch}{1.4}        
\setlength{\tabcolsep}{8pt}              
\begin{tabular}{lccccc}
\toprule
\textbf{Evaluator Setup} & \textbf{Acc.} & \textbf{Prec.} & \textbf{Recall} & \textbf{F1} & \textbf{Cohen's $\kappa$} \\
\midrule
Single LLM Evaluation (GPT-5) & 82.0 & 81.7 & 84.1 & 83.5 & 0.82 \\
Multi-LLM Arbitration (Ours)  & \textbf{89.6} & \textbf{90.4} & \textbf{88.1} & \textbf{87.7} & \textbf{0.87} \\
\midrule
Human-Human (Upper Bound)     & \textit{93.2} & \textit{94.1} & \textit{92.5} & \textit{93.2} & \textit{0.92} \\
\bottomrule
\end{tabular}
\end{table}

The multi-LLM panel raises Cohen's $\kappa$ from 0.82 (single GPT-5 judge) to 0.87, recovering most of the gap to the 0.92 ceiling set by inter-human agreement. We attribute this to three principled design choices: (1)~\textit{reference-grounded} evaluation against human-expert solutions rather than open-ended judgment, so that only demonstrable contradictions with established facts and theorems are flagged; (2)~\textit{root-cause focus} on path-independent logical errors (CIE / LAE / DRE), eliminating bias from varying reasoning styles across models; and (3)~\textit{absolute penalty} by error count ($P_i = V_i - \tau \sum_k x_{i,k}$) rather than step-level accuracy, removing the confound of differing output lengths.

\subsection{Stability Across Repetition, Judges, and Subsets}
\label{sec:appendix-judge-stability}

A reliable evaluator should not only correlate with human judgment on average; its rankings must also be stable across stochastic decoding, alternative judge panels, and different evaluation slices. We probe three dimensions on a 200-sample subset drawn from the Rigorous Process Set:

\begin{itemize}
    \item \textbf{Intra-judge:} The same judge evaluates each problem 3$\times$ with temperature $=0.6$. We report the agreement rate across the three runs.
    \item \textbf{Inter-judge:} Three independent judges (Claude-Sonnet-4.6, Gemini-3-Flash, GPT-5) evaluate the same subset; we report pairwise agreement aggregated across pairs.
    \item \textbf{Cross-subset ranking:} Model PES rankings are compared across slices defined by subject (4), question type (3), difficulty label (3), and 100 random half-splits over the full 2{,}124 questions. We report Kendall's $\tau$ (mean$\pm$std).
\end{itemize}

Table~\ref{tab:pes-stability} summarizes the results. Intra-judge repeatability exceeds 91\%, inter-judge consensus exceeds 84\%, and ranking consistency under all four subset partitions exceeds Kendall's $\tau = 0.9$. Together, these results indicate that the PES metric induces a stable model ordering that is robust to the specific evaluator instance and to the slice of LiveK12Bench used for evaluation.

\begin{table}[!htbp]
\centering
\caption{\textbf{Stability of process-error evaluation on a 200-sample slice of the Rigorous Process Set.} The arbitration panel is consistent across repeated runs, alternative judges, and different evaluation partitions.}
\label{tab:pes-stability}
{\small
\begin{tabular}{llcccc}
\toprule
Dimension & Metric & CIE & LAE & DRE & Overall \\
\midrule
Intra-judge & Agreement rate & 92.4\% & 95.5\% & 92.8\% & \textbf{91.7\%} \\
\midrule
Inter-judge & Agreement rate & 81.7\% & 85.9\% & 83.9\% & \textbf{84.3\%} \\
\midrule
\multirow{4}{*}{\shortstack[l]{Cross-subset\\Ranking\\(Kendall's $\tau$,\\mean$\pm$std)}}
    & By subject (4)     & \multicolumn{4}{c}{\textbf{0.901$\pm$0.025}} \\
    & By type (3)        & \multicolumn{4}{c}{\textbf{0.939$\pm$0.049}} \\
    & By difficulty (3)  & \multicolumn{4}{c}{\textbf{0.932$\pm$0.045}} \\
    & Random half        & \multicolumn{4}{c}{\textbf{0.914$\pm$0.038}} \\
\bottomrule
\end{tabular}}
\end{table}

\subsection{On the Use of Completion Tokens as the Efficiency Anchor}
\label{sec:appendix-token-anchor}

The reasoning-efficiency dimension of LiveK12Bench (ARL and $\text{Acc}_{\leq r}$) anchors on completion-token count rather than wall-clock time or FLOPs. We adopt this anchor for three reasons. First, each generated token corresponds to exactly one autoregressive forward pass and one billing unit, making tokens directly tied to both computational cost and serving latency; token-based metrics are also the established convention in the recent reasoning-efficiency literature~\cite{liu2025efficientreasoning,luo2025o1pruner}. Second, we measure API completion tokens (including ``thinking'' tokens where applicable), which is (i)~unaffected by Chain-of-Thought visibility, (ii)~faithful to real-world deployment because we use \emph{default} decoding parameters and each model's native tokenizer rather than artificial constraints, and (iii)~deliberately inclusive of verbosity, CoT style, and decoding strategy as \emph{part of} the efficiency profile being evaluated, since these factors directly impact inference latency and cost. Third, empirically, the residual tokenizer bias is small: prompt-token counts for identical inputs differ by less than 9\% across the tokenizers used in our study, an order of magnitude below the $\sim$5$\times$ completion-token gaps driven by reasoning strategies. Together these observations make completion-token count a sound and reproducible anchor for cross-model efficiency comparison in our setting.

\section{Data Illustrations and Reasoning Examples}
\label{sec:examples}

\subsection{Multi-Disciplinary Data Visualization}
\begin{figure}[p]
    \centering
    \begin{subfigure}{\textwidth}
        \centering
        \includegraphics[width=\textwidth, height=0.25\textheight, keepaspectratio]{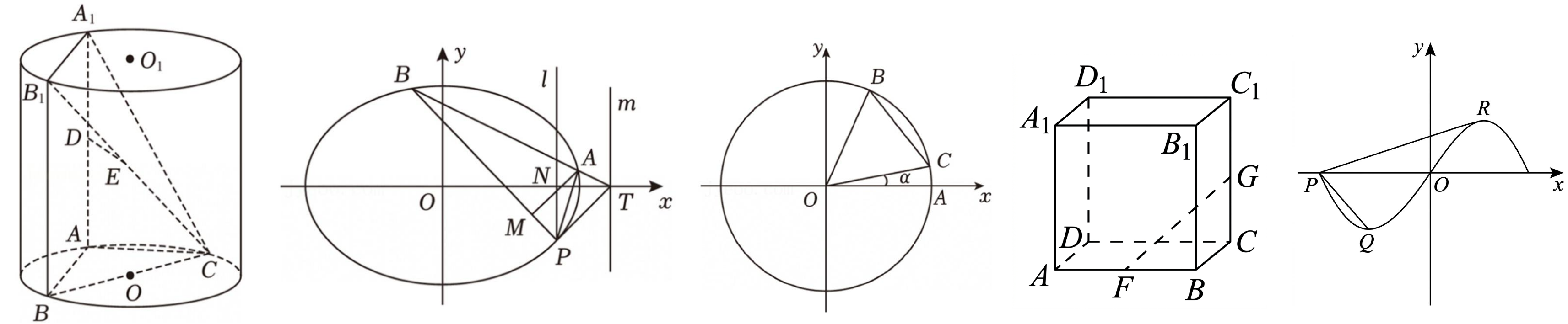}
        \caption{Representative examples of mathematical diagrams.}
    \end{subfigure}
    \vspace{0.5cm}
    \begin{subfigure}{\textwidth}
        \centering
        \includegraphics[width=\textwidth, height=0.25\textheight, keepaspectratio]{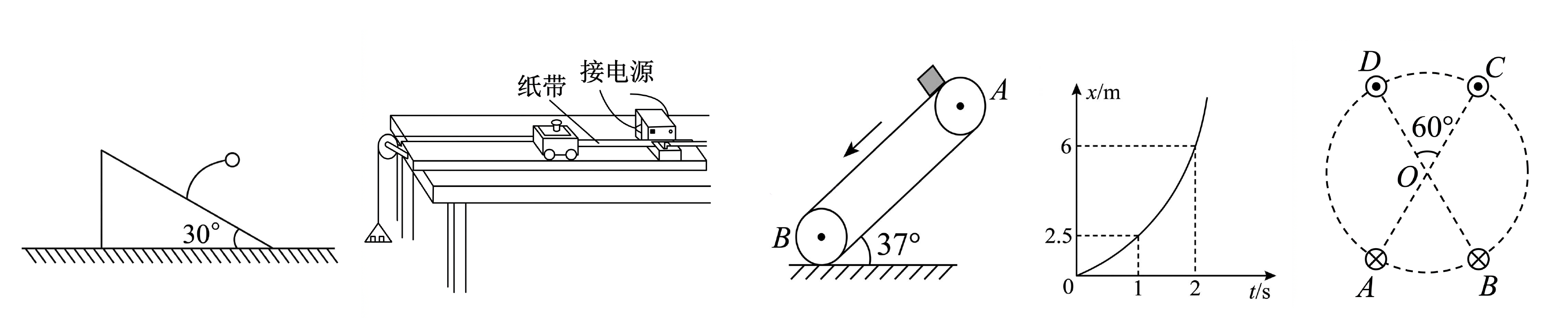}
        \caption{Representative examples of physics diagrams.}
    \end{subfigure}
    \vspace{0.5cm}
    \begin{subfigure}{\textwidth}
        \centering
        \includegraphics[width=\textwidth, height=0.25\textheight, keepaspectratio]{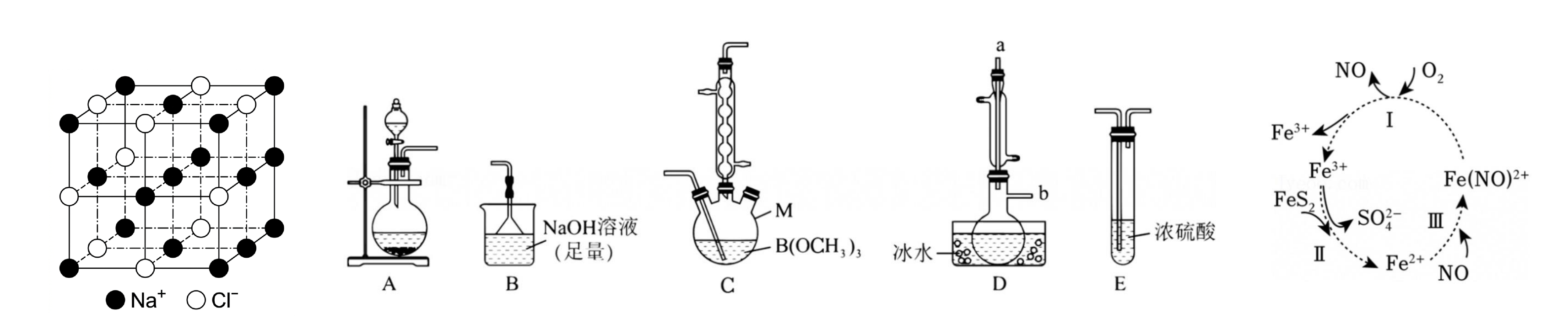}
        \caption{Representative examples of chemical diagrams.}
    \end{subfigure}
    \vspace{0.5cm}
    \begin{subfigure}{\textwidth}
        \centering
        \includegraphics[width=\textwidth, height=0.25\textheight, keepaspectratio]{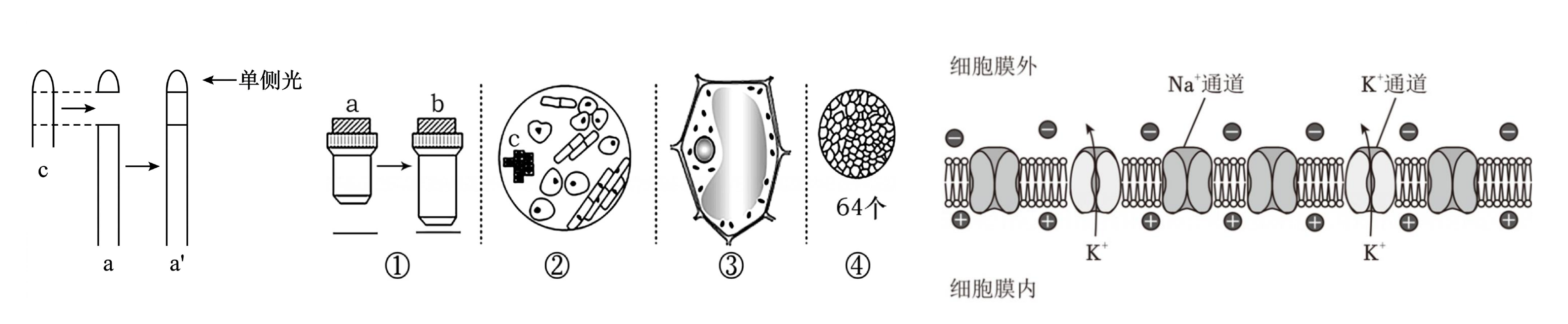}
        \caption{Representative examples of biological diagrams.}
    \end{subfigure}
\caption{\textbf{Multi-disciplinary data visualizations from the curated LikeK12Bench dataset.}}
\label{Figure:showcases}
\end{figure}

Herein, we present the elaborate visualization of the diverse scientific images contained in LiveK12Bench. As illustrated in Figure \ref{Figure:showcases}, the academic questions span 4 disciplines and 2,725 knowledge points, providing a comprehensive assessment of the sophisticated reasoning capabilities of MLLMs.

\subsection{Challenging Reasoning Responses}
Within LiveK12Bench, we incorporate three challenging subsets to assess the problem-solving capability on complex reasoning instances. As presented in Table \ref{Table:Set1}, \ref{Table:Set2}, \ref{Table:Set3}, we provide the comprehensive presentation of GPT-5’s reasoning responses on curated challenging subsets of complex problems.

\begin{table}[p]
    \centering
    \caption{\textbf{GPT-5’s Reasoning Responses on the Complex Layout Set.}}
    \label{Table:Set1}
    \begin{tabular}{p{0.9\textwidth}}
        \toprule
        \multicolumn{1}{c}{\textbf{Challenging Problems in the Complex Layout Set}} \\
        \hline
        \rule{0pt}{3ex}
        \begin{minipage}[t]{0.23\linewidth}
            \centering
            \vspace{-8pt}
            \includegraphics[width=\linewidth]{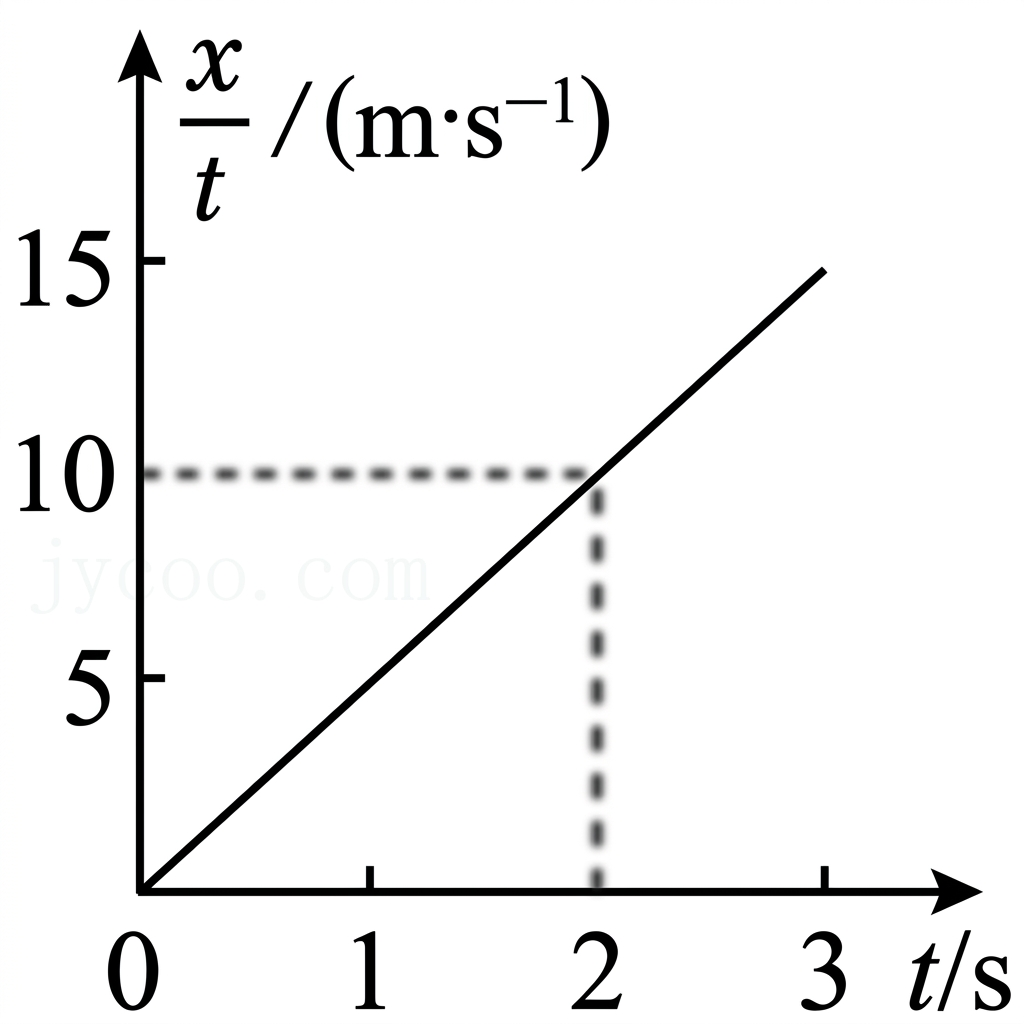}
        \end{minipage}%
        \hfill
        \begin{minipage}[t]{0.68\textwidth}
            \vspace{-5pt}
            \textbf{Problem.} The figure shows the relationship between the ratio $x/t$ and time $t$ for car A moving along a straight road. Which of the following statements is correct? \\
            A.\ The magnitude of car A’s acceleration is $5\,\mathrm{m/s^2}$. \\
            B.\ At $t=2\,\mathrm{s}$, the magnitude of car A’s velocity is $10\,\mathrm{m/s}$.
            C.\ Over $0\sim 2\,\mathrm{s}$, car A’s displacement is $10\,\mathrm{m}$. \\
            D.\ Over $0\sim 2\,\mathrm{s}$, the magnitude of displacement is $20\,\mathrm{m}$.
            \vspace{-1em}
        \end{minipage} 
        \vspace{1.5ex} \\
        \hline
        \rule{0pt}{3ex}
        \textbf{Reasoning.}
        \textit{Statement A:}
        Starting from the kinematic equation $x = v_{0} t + \frac{1}{2} a t^{2}$, we can rearrange to obtain that $\frac{x}{t} = v_{0} + \frac{1}{2} a t$. From the $\frac{x}{t}$–$t$ graph, we infer $v_{0} = 0$, $\frac{1}{2} a = \frac{10}{2}\,\mathrm{m/s^{2}}$, which yields $a = 10\,\mathrm{m/s^{2}}$. \\
        \textit{Statement B:}
        \hlred{According to the annotations on the velocity–time graph, the velocity of car A at $t=2\,\mathrm{s}$ is $10\,\mathrm{m/s}$. So statement B is correct.} \\
        \textit{Statements C and D:}
        From $x = v_{0} t + \frac{1}{2} a t^{2}$, the displacement of car A in the interval $0$–$2\,\mathrm{s}$ is $x_{2} = 10 + \frac{1}{2} \times 10 \times 2^{2}\,\mathrm{m} = 30\,\mathrm{m}$. Hence, both statement C and statement D are incorrect. \\
        \textbf{Answer.}
        The correct choice is $B$. 
        \raisebox{-0.1cm}{\includegraphics[width=0.4cm, height=0.4cm]{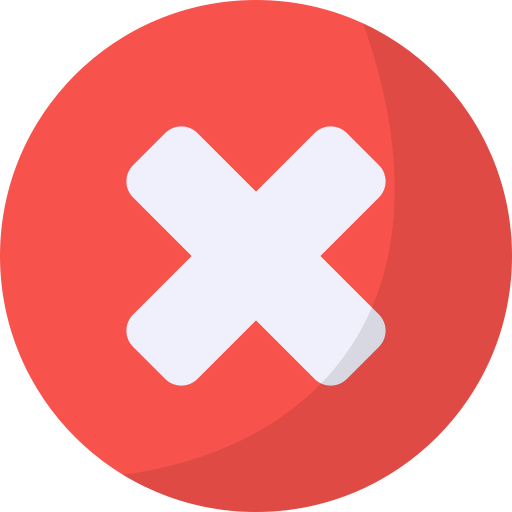}} \\
        \textbf{Standard Solution.}
        The correct answer is $D$.
        \raisebox{-0.1cm}{\includegraphics[width=0.4cm, height=0.4cm]{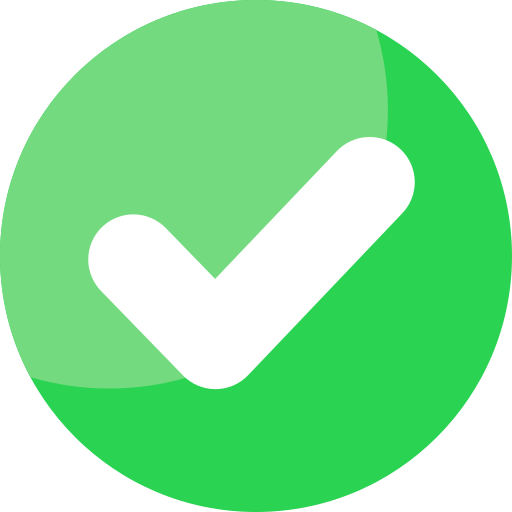}}
        \rule[-2ex]{0pt}{0pt} \\
        \hline
        \vspace{-1.5em}
        \begin{center}
            \includegraphics[width=0.9\linewidth]{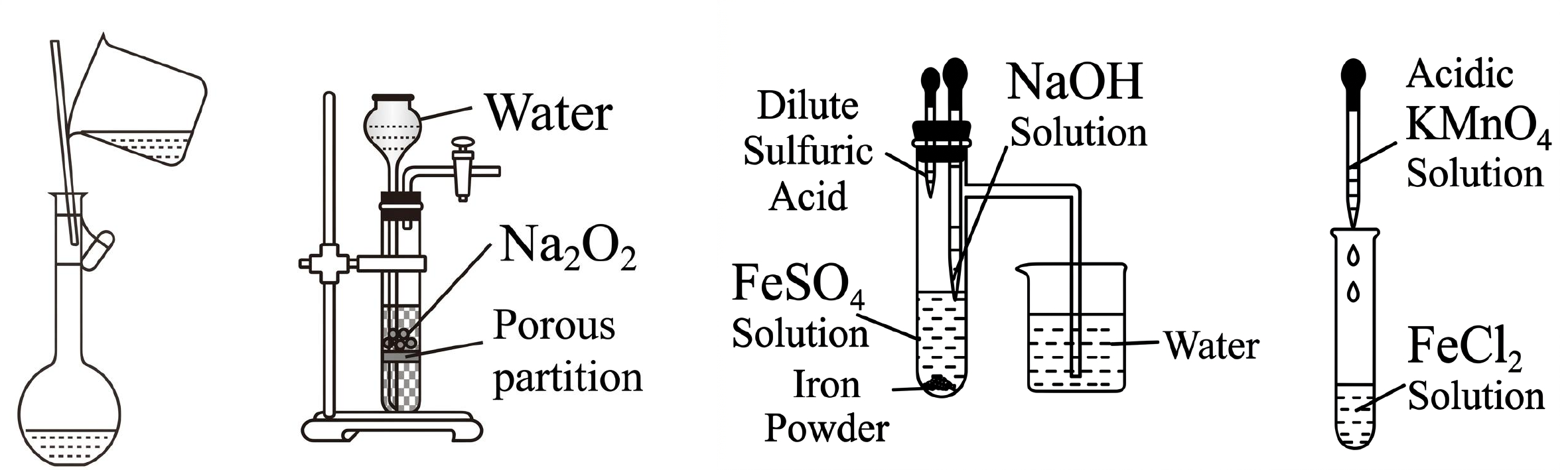}
        \end{center}
        \textbf{Problem.} Using the corresponding experiments with the apparatus shown, which of the following can achieve the intended experimental objective? \\
        A.\ Using Apparatus I to transfer a solution into a volumetric flask. \\
        B.\ Employing Apparatus II for the generation of $O_{2}$ via the reaction between $Na_{2}O_{2}$ and $H_{2}O$. \\
        C.\ Using Apparatus III to observe precipitate $Fe(OH)_{2}$ for an extended period. \\
        D.\ Utilizing Apparatus IV to demonstrate $Fe^{2+}$ exhibits reducing properties.
        \rule[-2ex]{0pt}{0pt} \\
        \hline
        \rule{0pt}{3ex}
        \textbf{Reasoning.}
        A.\ When transferring a solution into a volumetric flask, the tip of the glass rod should extend below the calibration line; therefore, A is incorrect.\\
        B.\ Sodium peroxide ($\mathrm{Na_2O_2}$) is a water-soluble powder; the apparatus in Figure 2 cannot be used to prepare a small amount of oxygen ($\mathrm{O_2}$) from the reaction between $\mathrm{Na_2O_2}$ and water ($\mathrm{H_2O}$); therefore, B is incorrect.\\
        C.\ \hlred{It is inappropriate to place dilute sulfuric acid in the same test tube. The acid will neutralize the added $\mathrm{OH^{-}}$ and dissolve $\mathrm{Fe(OH)_2}$.} \\
        D.\ The setup in Figure 4 demonstrates that $\mathrm{Fe^{2+}}$ has reducing properties.\\
        \textbf{Answer.}
        Therefore, the answer is $D$. 
        \raisebox{-0.1cm}{\includegraphics[width=0.4cm, height=0.4cm]{Figures/icons/no.png}} \\
        \textbf{Standard Solution.}
        Ultimately, option $C$ is the correct answer.
        \raisebox{-0.1cm}{\includegraphics[width=0.4cm, height=0.4cm]{Figures/icons/yes.png}}
        \rule[-2ex]{0pt}{0pt} \\
        \bottomrule
    \end{tabular}
\end{table}

\begin{table}[p]
    \centering
    \caption{\textbf{GPT-5’s Reasoning Responses on the Rigorous Process Set.}}
    \label{Table:Set2}
    \begin{tabular}{p{0.95\textwidth}}
        \toprule
        \multicolumn{1}{c}{\textbf{Challenging Problems in the Rigorous Process Set}} \\
        \hline
        \rule{0pt}{3ex}
        \textbf{Problem.} 
        Which of the following ionic equations for the specified reactions is correctly written? \\
        A. Reaction of chlorine with milk of lime to produce bleaching powder:
        $Cl_{2} + 2\,OH^{-} = Cl^{-} + ClO^{-} + H_{2}O$. \\
        B. Preparation of CO$_2$ using marble and dilute hydrochloric acid:
        $CO_{3}^{2-} + 2\,H^{+} = CO_{2}\uparrow + H_{2}O$. \\
        C. Dropwise addition of sulfuric acid to a Ba(OH)$_2$ solution:
        $H^{+} + OH^{-} = H_{2}O$. \\
        D. Oxidation of sulfurous acid by hydrogen peroxide:
        $H_{2}SO_{3} + H_{2}O_{2} = 2\,H^{+} + SO_{4}^{2-} + H_{2}O$.
        \rule[-2ex]{0pt}{0pt} \\
        \hline
        \rule{0pt}{3ex}
        \textbf{Reasoning.}
        \textit{Option A:}
        Chlorine reacts with limewater (aqueous calcium hydroxide) to form bleaching powder. The essential redox step in cold alkaline solution is $ \mathrm{Cl_2 + 2\,OH^- \rightarrow Cl^- + ClO^- + H_2O} $. Calcium ions are spectator ions in this medium; including $ \mathrm{Ca^{2+}} $ in the net ionic equation is inappropriate. Moreover, omitting the coefficient “2” before $ \mathrm{OH^-} $ breaks stoichiometric balance. Hence, the equation written as $ \mathrm{Cl_2 + Ca(OH)_2 = Ca^{2+} + Cl^- + ClO^- + H_2O} $ is not a correct net ionic representation (it contains a spectator and is stoichiometrically incomplete). Therefore, $A$ is incorrect. \\
        \textit{Option B:}
        Carbonate from marble ($ \mathrm{CaCO_3} $) reacts with acid to release carbon dioxide: $ \mathrm{CO_3^{2-} + 2\,H^+ \rightarrow CO_2\uparrow + H_2O} $. This is the proper net ionic equation. Writing $ \mathrm{CaCO_3 + 2\,H^+ = Ca^{2+} + CO_2\uparrow + H_2O} $ mixes the dissolution/precipitation step of calcium with the carbonate acidification, producing a total (not net) ionic form that unnecessarily includes the product $ \mathrm{Ca^{2+}} $ as a non-participating spectator in the acid–carbonate neutralization. Since the prompt calls for the ionic (net ionic) equation, the provided form is not the correct net ionic representation. Therefore, $B$ is incorrect. \\
        \textit{Option C:}
        Barium hydroxide solution reacts with sulfuric acid to yield insoluble barium sulfate and water. Representing both precipitation and neutralization steps, a correct ionic equation is $ \mathrm{Ba^{2+} + SO_4^{2-} + 2\,H^+ + 2\,OH^- \rightarrow BaSO_4\downarrow + 2\,H_2O} $. This expression conserves atoms and charge: on the left, total charge $ = (+2) + (-2) + (+2) + (-2) = 0 $; on the right, the precipitate is neutral and water is neutral. \hlred{Alternatively, one can decompose it into two canonical net ionic processes: $ \mathrm{Ba^{2+} + SO_4^{2-} \rightarrow BaSO_4\downarrow} $, $ \mathrm{H^+ + OH^- \rightarrow H_2O} $, and then sum them consistently. The provided ionic equation for $C$ is therefore correct.} \\
        \textit{Option D:}
        Oxidation of sulfite by hydrogen peroxide in aqueous solution is properly given (in neutral/alkaline media) by $ \mathrm{SO_3^{2-} + H_2O_2 \rightarrow SO_4^{2-} + H_2O} $. Writing $ \mathrm{H_2SO_3 + H_2O_2 = 2\,H^+ + SO_4^{2-} + H_2O} $ is problematic: \hlred{$ \mathrm{H_2SO_3} $ is not a stable isolable species in water (the system is better described in terms of $ \mathrm{SO_3^{2-}} / \mathrm{HSO_3^-} $), and introducing $ \mathrm{2\,H^+} $ in the products without specifying the medium leads to an inconsistent charge accounting relative to neutral reactants. Consequently, the stated ionic form is not the appropriate net ionic representation. Therefore, $D$ is incorrect.} \\
        \textbf{Answer.}
        Only the ionic equation in Option $C$ is correct. 
        \raisebox{-0.1cm}{\includegraphics[width=0.4cm, height=0.4cm]{Figures/icons/no.png}} \\
        \textbf{Standard Solution.}
        The correct choice is $D$.
        \raisebox{-0.1cm}{\includegraphics[width=0.4cm, height=0.4cm]{Figures/icons/yes.png}}
        \rule[-2ex]{0pt}{0pt} \\
        \bottomrule
    \end{tabular}
\end{table}

\begin{table}[p]
    \centering
    \caption{\textbf{GPT-5’s Reasoning Responses on the Long-Horizon Reasoning Set.}}
    \label{Table:Set3}
    \begin{tabular}{p{0.95\textwidth}}
        \toprule
        \multicolumn{1}{c}{\textbf{Challenging Problems in the Long-Horizon Reasoning Set}} \\
        \hline
        \rule{0pt}{3ex}
        \begin{minipage}[t]{0.25\linewidth}
            \centering
            \vspace{-8pt}
            \includegraphics[width=\linewidth]{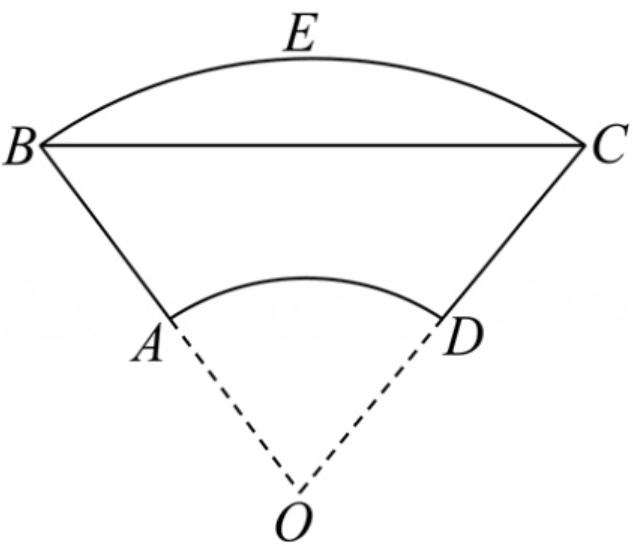}
        \end{minipage}%
        \hfill
        \begin{minipage}[t]{0.68\textwidth}
            \vspace{0pt}
            \textbf{Problem.} As illustrated, consider an annular sector-shaped flowerbed $ABCD$. The radius of the outer circular arc is twice the radius of the inner circular arc, the total perimeter is fixed at $2l$, and the central angle is $\alpha$ with $0<\alpha<\pi$. \\
            (1) When $\alpha=2$ (radians), determine the distance from the midpoint $E$ of the arc $\widehat{BC}$ to the chord $BC$. 
        \end{minipage} \\
        (2) For which value of $\alpha$ (in radians) does the area of the annular sector attain its maximum, and what is the maximum area?
        \rule[-2ex]{0pt}{0pt} \\
        \hline
        \rule{0pt}{3ex}
        \textbf{Reasoning.}
        (1) Consider the annular sector ABCD with inner radius r and outer radius 2r. Its perimeter is fixed and equal to $2l$, and its central angle is $2$ (in radians). Let the inner radius be $r$, so the two radial boundary segments each have length $2r - r = r$, and hence $AB = CD = OA = OD = r$. The arc lengths are $|AD| = r\alpha$ for the inner arc and $|BC| = 2r\alpha$ for the outer arc, where $\alpha$ denotes the angle parameter used below (in part (2)).
        Therefore, the perimeter constraint gives $r\alpha + 2r\alpha + 2r = 2l$, whence
        $r = \frac{2l}{3\alpha + 2}$.
        Let $OE$ intersect the chord $BC$ at $F$. By the perpendicular-from-center-to-chord theorem, we have $OE \perp BC$, and $\angle BOE = \tfrac{1}{2}\angle BOC = 1$. When $\alpha = 2$, the above relation yields $r = \frac{2l}{3\alpha + 2} = \frac{2l}{8} = \frac{l}{4}$. In the normalized case $l = 1$, this gives $r = \tfrac{1}{4}$. Since the distance from the center to the chord subtending angle $2$ equals $(\text{outer radius})\cdot\cos 1$, we obtain
        $OF = (2r)\cos 1 = \tfrac{1}{2}\cos 1$,
        and hence the distance from the point $E$ on the outer circle along $OE$ to the chord $BC$ is
        $EF = (2r) - OF = \tfrac{1}{2} - \tfrac{1}{2}\cos 1 = \tfrac{1}{2}(1 - \cos 1)$. \\
        (2) From (1) we have $r = \frac{2l}{3\alpha + 2}$. The area of the annular sector is the difference of the areas of the outer and inner sectors:
        $S_{\text{annulus}} = S_{OBC} - S_{OAD} = \tfrac{1}{2}(2r)^{2}\alpha - \tfrac{1}{2}r^{2}\alpha = \tfrac{3}{2}\alpha r^{2}$.
        Substituting $r = \frac{2l}{3\alpha + 2}$ gives
        $S_{\text{annulus}} = \frac{3}{2}\alpha\left(\frac{2l}{3\alpha + 2}\right)^{2} = \frac{6\alpha\,l^{2}}{(3\alpha + 2)^{2}} = \frac{6l^{2}}{9\alpha + 12 + \frac{4}{\alpha}}$.
        \hlred{By the AM–GM inequality ......} \\
        \textbf{Answer.}
        (1) The distance is $\tfrac{1}{2} - \tfrac{1}{2}\cos 1$. 
        \raisebox{-0.1cm}{\includegraphics[width=0.4cm, height=0.4cm]{Figures/icons/yes.png}} \\
        \hlred{(2) No answer is obtained owing to response-time constraints.} 
        \raisebox{-0.1cm}{\includegraphics[width=0.4cm, height=0.4cm]{Figures/icons/no.png}} \\
        \textbf{Standard Solution.}
        (1) $\frac{1}{2}(1-\cos 1)$;
        \raisebox{-0.1cm}{\includegraphics[width=0.4cm, height=0.4cm]{Figures/icons/yes.png}} \\
        (2) By the AM–GM inequality, $9\alpha + \frac{4}{\alpha} \ge 2\sqrt{9\alpha\cdot\frac{4}{\alpha}} = 12$, hence
        $S_{\text{annulus}} \le \frac{6l^{2}}{12 + 12} = \frac{l^{2}}{4}$,
        with equality if and only if $9\alpha = \frac{4}{\alpha}$, i.e., $\alpha = \frac{2}{3}$. Therefore, $S_{\text{annulus}}$ attains its maximum value $\frac{l^{2}}{4}$ precisely when $\alpha = \frac{2}{3}$.
        $\frac{2}{3}$, $\frac{1^{2}}{4}$.
        \raisebox{-0.1cm}{\includegraphics[width=0.4cm, height=0.4cm]{Figures/icons/yes.png}}
        \rule[-2ex]{0pt}{0pt} \\
        \bottomrule
    \end{tabular}
\end{table}

\section{Empirical Evidence for Data Contamination Mitigation}
\label{sec:appendix-contamination}

A central design goal of LiveK12Bench is to reduce data-contamination risk by continually ingesting examination papers \emph{newer than} the training cutoffs of mainstream LMMs. To verify that this design has the intended effect, we conduct a temporal-split experiment that compares model performance on two splits of LiveK12Bench drawn under identical sampling protocols and differing only in release date.

\textbf{Setup.} The 2025-06 split contains 358 questions sampled from exam papers released around June 2025, using the same per-subject and per-modality stratification as the main 2026-03 split reported in Table~\ref{tab:main_results}. The two splits therefore control everything except the release time of the source exams---an effective natural experiment on contamination, since LMMs cannot have trained on questions that postdate their training cutoffs.

\textbf{Results.} Table~\ref{tab:contamination} reports Overall Exam Score (OES) per subject on the 2025-06 split, with subscripts denoting the score difference relative to the 2026-03 split. Every model evaluated achieves a \emph{higher} OES on the older split, and the average gap $\overline{\Delta}$ is positive across all five models, ranging from $+3.0$ for Claude-opus-4.6 to $+7.0$ for Qwen3-VL-8B. The gap is consistent rather than concentrated in a single subject or model family, and it is largest for the smallest open-source model, which is consistent with the hypothesis that smaller models rely more heavily on memorization when available.

\begin{table}[!htbp]
\centering
\setlength{\tabcolsep}{5pt}
\caption{\textbf{Overall Exam Scores on the 2025-06 split of LiveK12Bench.} \textcolor{red}{Subscripts} report the score increase relative to the 2026-03 split (Table~\ref{tab:main_results}). The consistently positive gap $\overline{\Delta}$ supports the hypothesis that earlier exams are more susceptible to contamination, and that ingesting newer papers reduces this risk.}
\label{tab:contamination}
{\small
\begin{tabular}{lcccc|c}
\toprule
Model & Math & Physics & Chemistry & Biology & $\overline{\Delta}$ \\
\midrule
Gemini-3-pro     & $95.6_{\textcolor{red}{+5.3}}$  & $88.7_{\textcolor{red}{+0.8}}$ & $81.0_{\textcolor{red}{+4.3}}$ & $83.2_{\textcolor{red}{+4.6}}$  & \textbf{+3.7} \\
Claude-opus-4.6  & $95.0_{\textcolor{red}{+5.0}}$  & $87.4_{\textcolor{red}{+3.7}}$ & $73.5_{\textcolor{red}{+2.4}}$ & $75.0_{\textcolor{red}{+0.9}}$  & \textbf{+3.0} \\
GPT-5            & $93.4_{\textcolor{red}{+7.9}}$  & $76.2_{\textcolor{red}{+3.5}}$ & $49.6_{\textcolor{red}{+5.4}}$ & $55.5_{\textcolor{red}{+1.8}}$  & \textbf{+4.7} \\
Kimi-k2.5        & $95.9_{\textcolor{red}{+8.1}}$  & $88.7_{\textcolor{red}{+3.3}}$ & $79.5_{\textcolor{red}{+7.7}}$ & $73.9_{\textcolor{red}{+0.7}}$  & \textbf{+4.9} \\
Qwen3-VL-8B      & $87.5_{\textcolor{red}{+10.3}}$ & $62.3_{\textcolor{red}{+5.5}}$ & $43.5_{\textcolor{red}{+4.2}}$ & $53.4_{\textcolor{red}{+7.9}}$  & \textbf{+7.0} \\
\bottomrule
\end{tabular}}
\end{table}

\textbf{Discussion.} These results confirm the causal premise underlying the LiveK12Bench data pipeline: \emph{LMMs cannot learn from future exams.} The pipeline therefore offers a principled way to mitigate contamination risk by routinely refreshing the test set with newly released papers. We emphasize that this argument supports \emph{mitigation} rather than \emph{elimination}: some new questions may still be slight variants of pre-existing problems, and a residual contamination risk persists. The continually ingested newer splits are best interpreted as a moving frontier that bounds, rather than removes, this risk.

\end{document}